\pdfoutput=1

\documentclass[11pt]{article}

\usepackage[preprint]{acl}

\usepackage{times}
\usepackage{latexsym}
\usepackage{amsmath}
\usepackage{amssymb}
\usepackage{tabularx}
\usepackage{booktabs}
\usepackage{tikz}
\usepackage{listings}
\usepackage{pgfplots}
\pgfplotsset{compat=1.18}
\usetikzlibrary{matrix,positioning}
\usepackage{siunitx}
 \usepackage{float} 
 \usepackage{placeins}
\usepackage{titletoc}

\usepackage[T1]{fontenc}

\usepackage[utf8]{inputenc}

\usepackage{microtype}

\usepackage{inconsolata}

\usepackage{graphicx}

\usepackage{xcolor}
\usepackage{listings}

\definecolor{codegreen}{rgb}{0,0.6,0}
\definecolor{codegray}{rgb}{0.5,0.5,0.5}
\definecolor{codepurple}{rgb}{0.58,0,0.82}
\definecolor{backcolour}{rgb}{0.95,0.95,0.92}

\lstdefinestyle{mystyle}{
    backgroundcolor=\color{backcolour},
    commentstyle=\color{codegreen},
    keywordstyle=\color{magenta},
    numberstyle=\tiny\color{codegray},
    stringstyle=\color{codepurple},
    basicstyle=\ttfamily\footnotesize,
    breakatwhitespace=false,
    breaklines=true,
    captionpos=b,
    keepspaces=true,
    numbers=none, 
    numbersep=5pt,
    showspaces=false,
    showstringspaces=false,
    showtabs=false,
    tabsize=2,
    frame=single 
}

\lstdefinestyle{promptstyle}{
    basicstyle=\ttfamily\small,
    breaklines=true,
    frame=single,
    backgroundcolor=\color{white}, 
    aboveskip=1em,
    belowskip=1em,
    showstringspaces=false,
    columns=fullflexible,
    keepspaces=true
}

\title{Dissecting Physics Reasoning in Small Language Models:\\ A Multi-Dimensional Analysis from an Educational Perspective}

\author{Nicy Scaria\thanks{Equal contribution}, Silvester John Joseph Kennedy\footnotemark[1], Krishna Agarwal, \\
\textbf{Diksha Seth, Deepak Subramani}\\
Computational and Data Sciences, Indian Institute of Science, India \\
\texttt{nicyscaria@iisc.ac.in}}

\begin{document}
\maketitle
\begin{abstract}

Small Language Models (SLMs) offer privacy and efficiency for educational deployment, yet their utility depends on reliable multi-step reasoning. Existing benchmarks often prioritize final-answer accuracy, obscuring `right answer, wrong procedure' failures that can reinforce student misconceptions. This work investigates SLM physics reasoning reliability, stage-wise failure modes, and robustness under paired contextual rewrites. We introduce \textit{PhysBench}, comprising 3,162 high school and AP-level physics questions derived from OpenStax in a structured reference-solution format with Bloom’s Taxonomy annotations, plus 2,700 paired culturally contextualized variants. Using P-REFS, a stage-wise evaluation rubric, we assess 10 SLMs across 58,000 responses. Results reveal a substantial reliability gap: among final-answer-correct solutions, 75-98\% contain at least one reasoning error. Failure modes shift with model capability; weaker models fail primarily at interpretation/modeling, while stronger models more often fail during execution. Paired contextual variations have a minimal impact on top models but degrade the performance of mid-tier models. These findings demonstrate that safe educational AI requires evaluation paradigms that prioritize reasoning fidelity over final-answer correctness.
\end{abstract}

\section{Introduction}\label{introduction}

\begin{figure*}[h]
\centering
\includegraphics[width=0.8\textwidth]{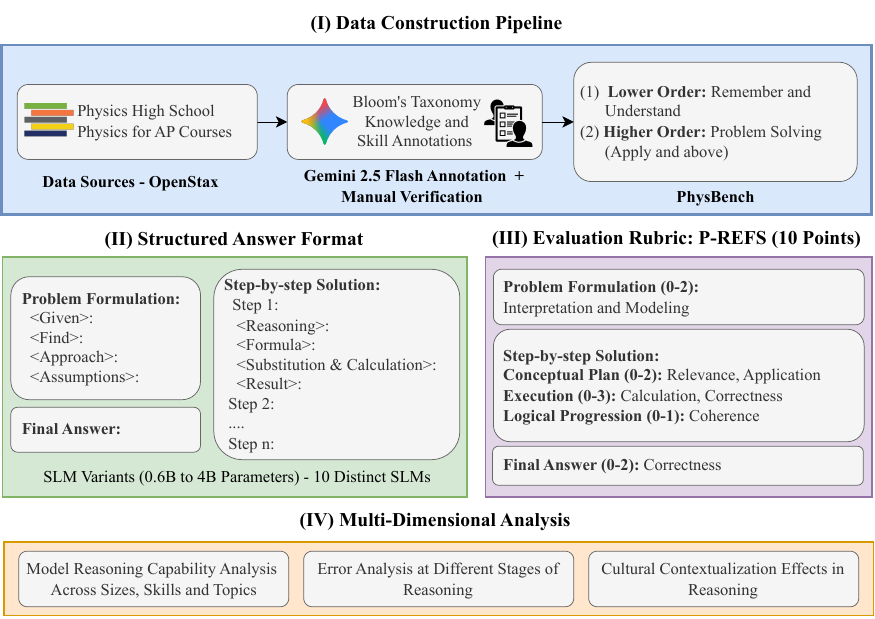}
\caption{Multi-dimensional physics reasoning evaluation pipeline using OpenStax problems with structured answers and P-REFS (10-point rubric) for analyzing performance, errors, and contextualization effects.}
\label{block_diagram}
\end{figure*}

Small Language Models (SLMs) are becoming increasingly practical for deployment in settings constrained by cost, latency, and privacy, making them attractive for educational support including on-device tutoring \cite{sun2020mobilebert,phi3,wei2025small}. However, educational usefulness depends on reasoning reliability, not only final-answer accuracy: solutions that reach correct answers via incorrect setups, misuse of principles, or faulty calculations can reinforce wrong procedures \cite{lanham2023measuring,lee2025evaluating,xia2025evaluating}. Physics provides a stringent testbed because it demands accurate scenario interpretation, appropriate physical modeling, and procedurally correct multi-step computation with unit consistency.

Most evaluations in quantitative domains emphasize final-answer correctness and provide limited visibility into where reasoning fails \cite{srivastava2025towards}. To make SLM evaluation actionable, we need diagnostics that separate early-stage failures (interpretation/modeling) from late-stage failures (calculation and unit consistency), characterize how these vary across topics and cognitive demands, and assess robustness under culturally localized rewrites. Educational problems are often rewritten into culturally familiar scenarios to improve engagement \cite{cordova1996intrinsic}. Models that genuinely understand the underlying physics should maintain stable reasoning under such contextual rewrites, rather than relying on brittle pattern matching to the original phrasing.

Figure~\ref{block_diagram} summarizes our approach. We introduce \textit{PhysBench}, comprising 3,162 high school and AP-level physics questions derived from OpenStax \cite{openstax,ap_physics} in a structured solution format with Bloom's Taxonomy annotations, plus 2,700 contextualized variants across three world regions. We evaluate 10 SLMs using P-REFS (Physics Reasoning Evaluation Framework for Step-by-step Analysis), a stage-wise rubric operationalized via an LLM-as-a-judge pipeline \cite{chiang2023can,thakur2024judging}. Across over 58,000 responses, we find a substantial reasoning reliability gap: among final-answer-correct solutions, 75-98\% still contain at least one error, with failure modes shifting from early-stage interpretation/modeling errors in weaker models to late-stage calculation and unit-consistency errors in stronger models. These diagnostics inform both educational deployment decisions and targeted improvements in model training, motivating interventions such as formulation-focused training and reasoning verification.

\paragraph{Contributions:} We introduce (i) \textit{PhysBench}: 3,162 high school and AP-level physics questions with structured solutions and Bloom's Taxonomy annotations, plus 2,700 culturally contextualized variants across three regions; (ii) P-REFS: a stage-wise rubric evaluating problem formulation, conceptual planning, procedural execution, and final answers; and (iii) multi-dimensional diagnostics isolating reasoning breakdown across topics, cognitive demands, and cultural contexts.

\section{Related Work}

Numerous benchmarks evaluate LLM performance on high school-level physics questions, spanning broad academic evaluations that include physics as a subset (e.g., MMLU \cite{hendrycks2020measuring}, C-Eval-STEM \cite{huang2023c}), science-oriented QA benchmarks (e.g., ScienceQA \cite{lu2022learn}, SciBench \cite{wang2023scibench}), and physics-focused suites (e.g. PhyQA \cite{anand2023sciphyrag}). While these benchmarks enable progress, many emphasize final-answer accuracy and provide limited insight into failures in multi-step quantitative reasoning. Moreover, few explicitly separate performance by cognitive demand or knowledge type, making it difficult to distinguish factual recall from procedural competence. Related benchmarks, such as JEEBench \cite{arora2023have}, PhysReason \cite{zhang2025physreason}, and OlympiadBench \cite{he2024olympiadbench}, target competition-style problems that are more complex than those in classroom physics.

Research shows that presenting content through familiar cultural frameworks, such as local names, settings, and practices, enhances engagement and learning in quantitative subjects \cite{marchee2022effectiveness, leinonen2021exploring, samo2018culture}. For SLMs to support diverse student populations effectively, they must maintain reasoning quality across both Western/generic and culturally familiar contexts; deterioration with cultural adaptation limits culturally responsive AI pedagogy. However, NLP studies report performance drops when prompts deviate from Western-centric assumptions \cite{ramezani2023knowledge, mushtaq2025worldview, veselovsky2025localized}, and controlled evaluations of cultural contextualization on quantitative problem solving remain limited for SLMs \cite{karim2025lost, tomar2025mathematics}, often relying on shallow named-entity substitutions that fail to embed grounded scenarios while preserving underlying physics and mathematical structure.

Prior work has shown that answer-only metrics can mask reasoning errors in quantitative domains \cite{lanham2023measuring, lee2025evaluating, xia2025evaluating}. In response, process-based evaluations and stepwise scoring protocols have emerged \cite{mirzadeh2024gsm, zhang2025physreason}, though many require strict answer formats that limit applicability across diverse model outputs, and many physics and mathematics benchmarks still default to final-answer grading or coarse correctness checks \cite{shojaee2025illusion, wang2025physunibench, fang2025mathodyssey}. PHYSICS EVAL \cite{physicseval} employs similar stepwise evaluation but focuses 
on mid-to-large LLMs. Our work complements this line by focusing on sub-4B SLMs and by analyzing robustness under culturally contextualized variants. Human grading provides the most reliable assessment but does not scale to large model-by-question evaluations, while rule-based parsers offer precision but struggle with diverse response formats. LLM-as-a-Judge frameworks provide a practical middle ground, with several studies reporting strong alignment between LLM judgments and human assessments \cite{chiang2023can, thakur2024judging, judgesurvey, llmsurvey}; however, systematic applications to fine-grained SLM reasoning in physics remain sparse. Recent work on evaluator robustness (e.g., JudgeBench \cite{judgebench}) highlights sensitivity to judge choice, suggesting judge-validation checks in rubric-based pipelines.

\section{Methodology}

\subsection{Model selection}\label{model_selection}

For this study, SLMs are defined as models with fewer than 4 billion parameters. The selected models included instruction-tuned variants of Llama3.2 (1B and 3B) \cite{llama3}, Phi4 Mini (3.84B) and its reasoning-focused variant \cite{phi4,phi4reasoning}, Gemma3 (1B and 4B) \cite{gemma3}, Qwen2.5 (1.5B) and Qwen2.5 DeepSeek Distil (1.5B) \cite{qwen2_5,deepseek}, and Qwen3 (0.6B and 1.7B) \cite{qwen3}. This set spans state-of-the-art SLM families and captures diversity in architecture, training methodology, and reasoning specialization. Including multiple sizes within families (e.g., Llama3.2, Gemma3, Qwen3) enables analysis of parameter scaling, while comparisons between standard and reasoning-optimized models (e.g., Phi4 Mini vs. Reasoning, Qwen2.5 vs. Qwen3) reveal the impact of reasoning-oriented training.

\subsection{Dataset creation}

The dataset was derived from OpenStax textbooks: High School Physics \cite{openstax} and Physics for AP{\textregistered} Courses \cite{ap_physics} (data preparation in Appendix~\ref{data_appendix}). To ensure self-contained problems, items requiring visual interpretation were excluded unless textual descriptions were sufficient. Each question was reformatted into a structured answer template (Figure~\ref{structured_format}) to enable full-solution evaluation. The final \textit{PhysBench} dataset comprised 3,162 questions spanning two question types: (1) \textbf{conceptual} (572 questions) focusing on theoretical understanding, and (2) \textbf{problem-solving} (2,590 questions) requiring numerical calculations and multi-step mathematical reasoning.

Each problem was annotated along the cognitive and knowledge dimensions of Bloom’s Taxonomy, distinguishing lower-order (Remember, Understand) and higher-order (Apply, Analyze, Evaluate, Create) reasoning, as well as factual, conceptual, and procedural knowledge categories. Annotations were generated using Gemini 2.5 Flash and manually verified for accuracy and consistency. The alignment between LLM-generated annotations and expert verification, along with the distribution of annotated data, is discussed in Appendix~\ref{blooms_appendix}.

\subsection{Contextualization of dataset}

To investigate how cultural contextualization affects SLM, we create culturally adapted variants of \textit{PhysBench} by rewriting problems with region-specific cultural and geographic elements while preserving the underlying physics. This enables controlled comparison of model reasoning across contextualized versions of the same problems.

\begin{figure}[h]
\centering
\includegraphics[width=0.45\textwidth]{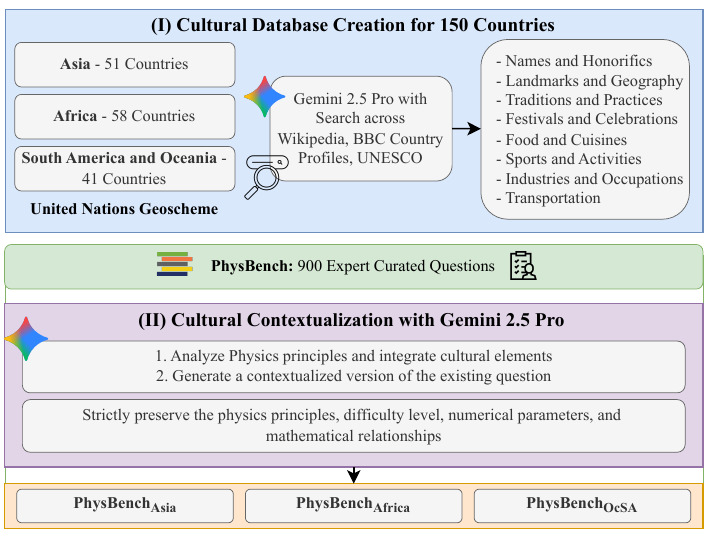}
\caption{Overview of the two-stage process for creating culturally contextualized physics benchmarks.}
\label{cultural_context}
\end{figure}

As illustrated in Figure~\ref{cultural_context}, the cultural contextualization process consists of two main stages. First, regional cultural databases were developed for 150 countries across Asia (51), Africa (58), and South America/Oceania (41), following the UN Geoscheme. Second, 900 questions were selected from \textit{PhysBench} and adapted using Gemini 2.5 Pro. Initial experiments with Gemini 2.5 Flash occasionally produced physically infeasible scenarios; Pro demonstrated superior capability in maintaining both cultural authenticity and physics validity. The model was instructed to embed cultural elements while strictly preserving the original physics principles, difficulty level, numerical parameters, and mathematical relationships. To ensure physical consistency, we generated multiple candidate contextualizations per question and selected the best through self-consistency checks. This produced three region-specific datasets: \textit{PhysBench}$_{\text{Asia}}$, \textit{PhysBench}$_{\text{Africa}}$, and \textit{PhysBench}$_{\text{OcSA}}$, each containing 900 problem solving questions. Details on the contextualization process are provided in Appendix~\ref{context_appendix}.

\subsection{Model inference} 

We conducted inference for all models in Section~\ref{model_selection} to produce a structured solution (Problem Formulation, Step-by-step Solution, Final Answer). We evaluate on \textit{PhysBench} (3,162 questions) and three culturally adapted subsets (\textit{PhysBench}$_{\text{Asia}}$, \textit{PhysBench}$_{\text{Africa}}$, \textit{PhysBench}$_{\text{OcSA}}$; 900 questions each) to measure baseline performance, Bloom's taxonomy and topic breakdowns, contextualization effects, and stage-wise error patterns. Inference and evaluation details (decoding settings, completion-token statistics) are in Appendix~\ref{inference_details}.

\subsection{Model evaluation}

All responses were evaluated using a structured rubric assessing three components: (1) Problem Formulation, (2) Step-by-step Solution, and (3) Final Answer. Each step was scored independently to enable fine-grained reasoning diagnostics.

\begin{table}[h]
\centering
\scriptsize
\caption{P-REFS: Physics Reasoning Evaluation Framework for Step-by-step Analysis (10 points)}
\label{tab:prefs}
\setlength{\tabcolsep}{2pt} 
\begin{tabular}{@{}ll@{}} 
\toprule
\textbf{Dimension} & \textbf{Evaluation Criteria}\\ 
\midrule
\multicolumn{2}{c}{\textbf{Problem Formulation (0-2 points)}} \\
\midrule
Interpretation & Are the given quantities and constraints correctly identified?\\
Modeling & Are the chosen approach and assumptions appropriate?\\
\midrule
\multicolumn{2}{c}{\textbf{Step-by-step Solution (0-6 points)}} \\
\midrule
\multicolumn{2}{c}{\textit{Conceptual Plan} (0-2 points)} \\
Relevance & Is the physics concept appropriate for this step? (0-1)\\
Application & Is the concept correctly applied? (0-1) \\
\multicolumn{2}{c}{\textit{Execution (0-3 points)}} \\
Calculation & Are all mathematical operations performed correctly? (0-1)\\
Correctness & Are results accurate (correct numerical values and units)? (0-2)\\
\multicolumn{2}{c}{\textit{Logical Progression (0-1 point)}} \\
Coherence & Does each step follow logically from the previous? (0-1)\\
\midrule
\multicolumn{2}{c}{\textbf{Final Answer (0-2 points)}} \\
\midrule
Correctness & Are results accurate, with correct numerical values and units?\\
\multicolumn{2}{l}{\scriptsize\textit{(Note: For MCQ, the selected option must exactly match the correct option)}} \\
\bottomrule
\end{tabular}
\end{table}

\paragraph{Problem-Solving Questions:} Performance was assessed using the P-REFS (Physics Reasoning Evaluation Framework for Step-by-step Analysis) rubric (Table~\ref{tab:prefs}), a 10-point framework evaluating three components: Problem Formulation (0-2 pts) for interpretation and modeling, Step-by-step Solution (0-6 pts) encompassing conceptual planning, execution accuracy, and logical progression, and Final Answer (0-2 pts) for correctness and completeness. For MCQs, full credit required an exact match with the correct option. Complete rubric specifications are detailed in Table~\ref{tab:prefs}.

\paragraph{Conceptual Questions:} For questions without numerical calculations, a simplified 5-point rubric evaluated Problem Formulation (0-1), Solution Execution (0-3), and Final Answer (0-1), with detailed criteria in Appendix~\ref{eval_appendix}.

To operationalize this evaluation framework, we implemented an LLM judge with specialized prompts for each question type. To ensure fair assessment regardless of output formatting, our evaluation protocol adapts to both template-compliant and free-form responses: structured solutions are scored part-by-part, while non-compliant responses are evaluated together in a single inference (Appendix~\ref{evaluation_protocol}). We conducted a comparative validation study of three candidate judges: Gemini 2.5 Flash~\citep{team2023gemini}, Qwen 3 32B~\citep{qwen3}, and Llama 3.3 70B~\citep{llama3}, selected for their strong performance on structured reasoning tasks~\citep{j1reinforcement2025, codejudgebench2025, thakur2024judging} and practical deployment feasibility for evaluating over 50,000 model responses.

Two independent expert reviewers with graduate-level qualifications in physics evaluated 550 question-response pairs per judge across all physics topics and knowledge and cognitive levels. Gemini 2.5 Flash which demonstrated the strongest alignment with expert judgments and lowest error rates, was selected as the primary evaluator. Following this validation, we conducted the evaluation of all responses in the \textit{PhysBench} benchmark using Gemini 2.5 Flash. Detailed validation methodology, inter-annotator agreement metrics and evaluator comparison are provided in Appendix~\ref{judge_validation}.

\begin{figure*}[h]
\centering
\includegraphics[width=0.9\textwidth]{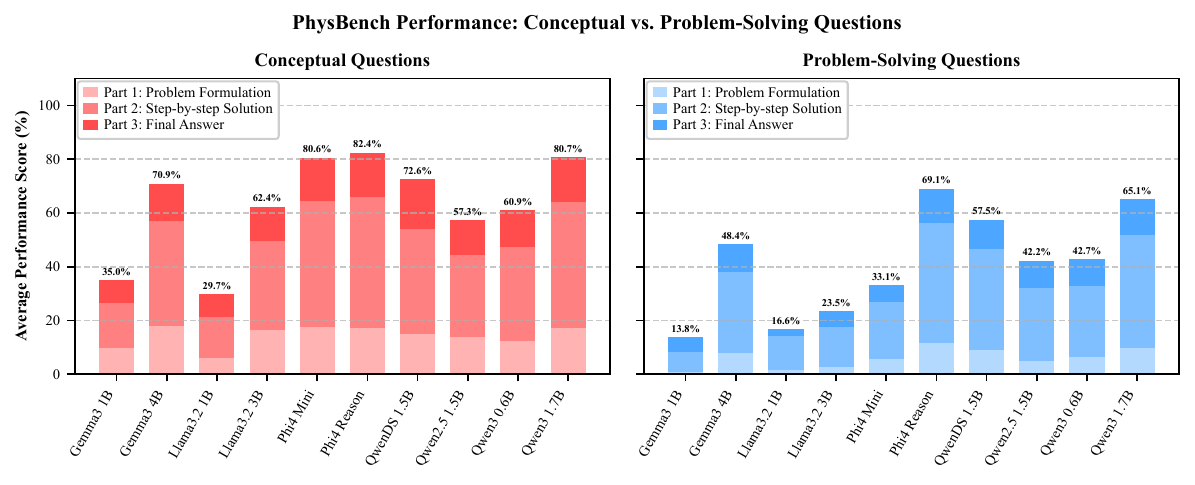}
\caption{Part-wise performance breakdown for \textit{PhysBench}. The height represents the average overall score (\%).}
\label{graph_1}
\end{figure*}

\subsection{Error analysis}

Using the structured P-REFS rubric, each incorrect or partially correct response was analyzed to identify the stage at which reasoning broke down. Errors were categorized into eight types: \textit{Interpretation}, \textit{Modeling}, \textit{Conceptual}, \textit{Concept Application}, \textit{Calculation}, \textit{Correctness}, \textit{Coherence}, and \textit{Final Answer}. The detailed definitions and illustrative examples provided in Appendix~\ref{appen_error_analysis}.

\section{Results and Analysis}

Our findings address all research questions, revealing patterns in SLMs' physics reasoning capabilities across question types, topics, knowledge, and cognitive demands, and cultural contexts. 

\subsection{Reasoning reliability by question type}

We first compare SLM performance on conceptual versus problem-solving questions (Figure~\ref{graph_1}), revealing a stark gap between the two question types.

\paragraph{Overall performance:} In conceptual questions (Figure~\ref{graph_1}, left), several models achieve high overall rubric scores, with non-trivial score contributions distributed across Problem Formulation and Step-by-step Solution. In particular, Phi4 Mini, Phi4 Reason and Qwen3 1.7B exceed 80\% overall, indicating that top-performing SLMs possess a strong baseline performance in conceptual reasoning.

In problem-solving questions (Figure~\ref{graph_1}, right), performance degrades sharply across all model families. The reduction is accompanied by visibly smaller contributions from both formulation and execution, suggesting that errors emerge early (problem setup) and persist through multi-step solution procedures. For example, Phi4 Mini drops from 80.6\% (conceptual) to 33.1\% (problem-solving), while Phi4 Reason drops from 82.4\% to 69.1\%, indicating that reasoning-optimized training improves robustness but does not eliminate the procedural bottleneck in multi-step quantitative physics. To ensure this gap is not an artifact of output structure or parsing of the three parts, we analyze template compliance and compare scores for template-compliant vs. non-compliant responses; compliance is not significantly correlated with P-REFS performance (Appendix~\ref{template_compliance}).

\begin{figure*}[h]
\centering
\includegraphics[width=0.9\textwidth]{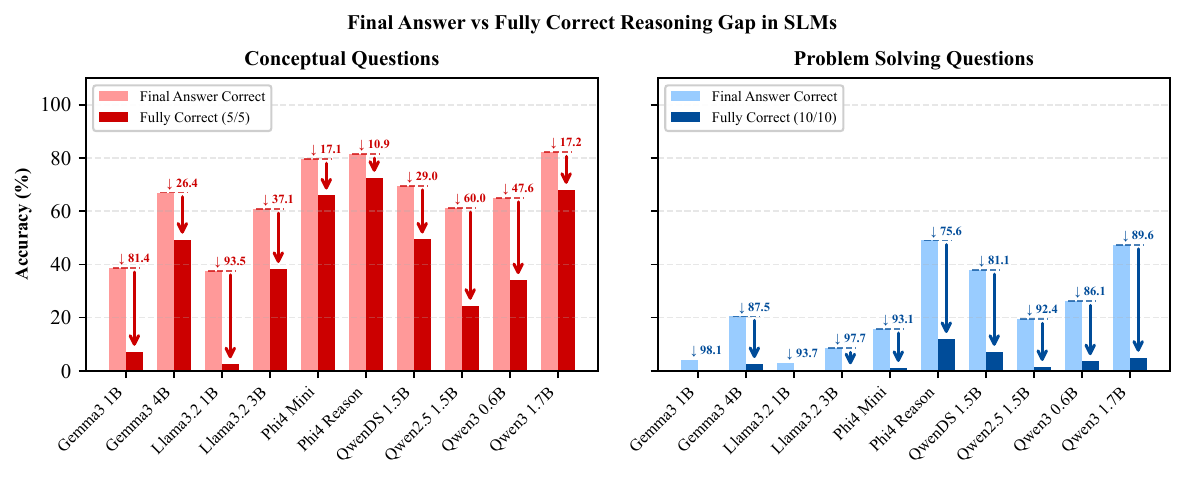}
\caption{Final answer vs. full correctness for conceptual and problem-solving questions. Gap percentages show correct answers with reasoning errors.}
\label{graph_2}
\end{figure*}

\paragraph{Reasoning performance gap:} The analysis of SLM performance reveals a consistent and significant discrepancy between the ability to arrive at the correct final answer and the capacity to generate valid reasoning chains. Figure~\ref{graph_2} illustrates this reasoning gap across both conceptual and problem-solving questions. The relative gap percentages indicate the proportion of correctly answered questions that contain errors in problem formulation, step-by-step execution, or both, revealing that SLMs frequently suffer from an illusion of competence in which the correct answers often stem from flawed problem setup, incorrect execution steps, or pattern matching rather than genuine physical understanding.

This discrepancy is most severe in problem-solving. As shown in the right panel of Figure~\ref{graph_2}, the gaps are catastrophic across all model families. For instance, Gemma3 4B exhibits an 87.5\% gap, meaning that nearly 9 out of 10 correct answers are accompanied by errors in formulation, execution, or both. Llama3.2 3B shows an even more extreme pattern; despite having very low answer accuracy, its gap is 97.7\%, indicating that nearly all correct answers are accompanied by incorrect problem formulation or execution steps. The pattern holds even for the strongest models: Qwen3 1.7B exhibits an 89.6\% gap, while Phi4 Reason, the best-performing model on problem-solving tasks, still demonstrates a 75.6\% gap, confirming that multi-step problem-solving remains a critical bottleneck for current SLMs. To test whether the gap is driven by problem setup, we run a formulation-scaffolded ablation that prepends the reference problem formulation. Full details and results are reported in Appendix~\ref{with_problemformulation}.

In contrast, SLMs are relatively more robust in conceptual reasoning. While gaps persist, they are significantly narrower than in problem solving. Phi4 Reason demonstrates better stability with a gap of only 10.9\%, indicating that it successfully produces correct formulations and explanations in approximately 9 out of 10 correctly answered cases. However, standard models still struggle; for example, Llama3.2 1B exhibits a 93.5\% gap, and Qwen2.5 1.5B shows a 60.0\% gap. 

\paragraph{Model scale, family, and reasoning-oriented training:}
Figures~\ref{graph_1} and~\ref{graph_2} also highlight systematic effects of model scale and training. Within model families, larger variants generally improve final-answer correctness and average rubric scores (e.g., Gemma 4B vs.\ 1B; Llama 3B vs.\ 1B; Qwen 3 1.7B vs.\ 0.6B), yet the final answer-fully correct reasoning reliability gap remains large in problem-solving, indicating that scaling alone does not ensure end-to-end procedural validity. Cross-family differences are also pronounced, suggesting that architecture and training recipe meaningfully influence both answer accuracy and solution reliability beyond parameter count. Finally, the reasoning-optimized Phi4 Reason consistently outperforms its non-reasoning counterpart on both overall scores and end-to-end correctness, and exhibits a smaller gap, implying that reasoning-focused training improves robustness; however, the persistence of substantial gaps even for the strongest models confirms that reliable multi-step quantitative reasoning remains an open challenge for current SLMs.

\subsection{Multi-dimensional performance: Physics domains and Bloom’s taxonomy}

Detailed breakdowns by physics domain and Bloom’s Taxonomy dimensions are provided in Appendix Figures \ref{fig:physics_topics_conceptual}--\ref{fig:knowledge_dimensions}; here we summarize the most consistent trends.

\paragraph{Physics domains:} Conceptual performance is relatively stable across physics topics for strong models, though some domains show variation in specific components. Domains requiring abstract reasoning (e.g., Electricity and Magnetism, Modern Physics) show less consistent Part 3 (Final Answer) and overall scores, while foundational topics show more uniform performance. In problem-solving, topic-specific differences are minor compared to the main problem: weak multi-step reasoning. Most models score poorly on Parts 1–2 across all domains, with only occasional exceptions. This shows that the primary bottleneck is the problem formulation and multi-step procedural execution, rather than a lack of domain-specific knowledge (Appendix Figures~\ref{fig:physics_topics_conceptual} and~\ref{fig:physics_topics_problemsolving}).

\begin{figure*}[t]
    \centering
    \includegraphics[width=0.8\textwidth]{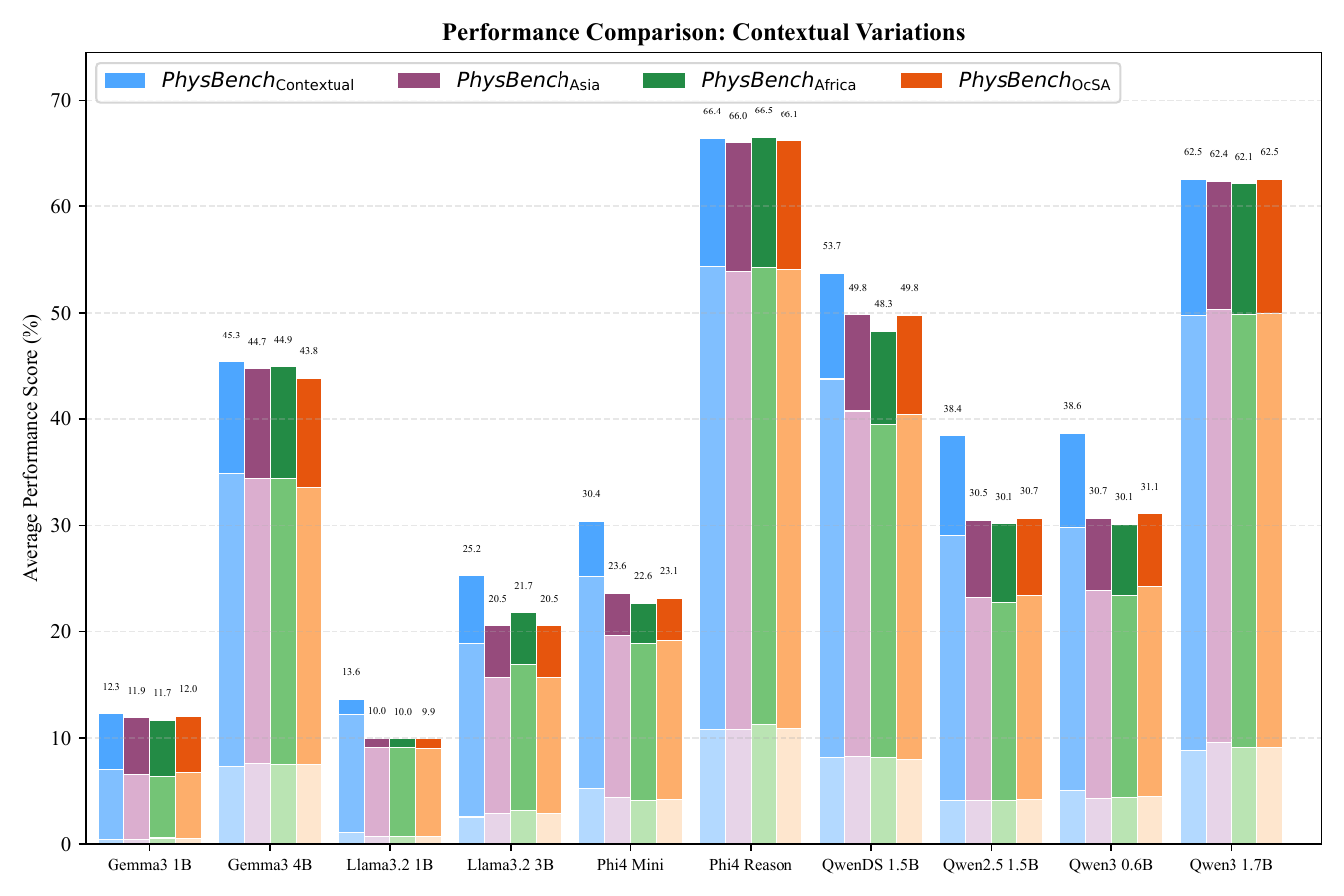}
    \caption{Performance across cultural contexts across \textit{PhysBench}$_{\text{Contextual}}$ (900-question baseline subset), \textit{PhysBench}$_{\text{Asia}}$, \textit{PhysBench}$_{\text{Africa}}$, and \textit{PhysBench}$_{\text{OcSA}}$.}
    \label{fig:cultural_contexts}
\end{figure*}

\paragraph{Cognitive dimension:} Performance shows a clear monotonic decline with increasing cognitive complexity. Models perform best on lower-order skills (Remember, Understand) and degrade substantially as tasks shift toward Apply/Analyze and especially Evaluate/Create. This degradation is visible across Problem Formulation and Step-by-step Solution, indicating that higher-order reasoning demands systematically weaken both formulation and procedural execution. The pattern holds across model families, confirming that cognitive complexity amplifies reasoning fragility regardless of model architecture (Appendix Figure~\ref{fig:cognitive_dimensions}).

\paragraph{Knowledge dimension:} Performance shows a clear hierarchy across knowledge types: models perform strongest on Factual knowledge, followed by Conceptual knowledge, and weakest on Procedural knowledge. This degradation is particularly pronounced in execution, indicating that systematic method application poses a greater challenge than factual recall or conceptual understanding. The pattern holds across model families, confirming that procedural execution is a core limiter for SLM physics competence (Appendix Figure~\ref{fig:knowledge_dimensions}).

\subsection{Performance across contextual variations}

We evaluate whether SLM performance remains stable when identical physics problems are presented in contextualized scenarios. Figure~\ref{fig:cultural_contexts} compares overall performance on the curated contextual subset (baseline) and its region-specific variants (Asia, Africa, and Oceania/South America).

The strongest models exhibit near-invariant performance. Phi4 Reason maintains scores between 66.0\% and 66.5\% across all regions (a variation of just 0.5\%), and Qwen3 1.7B remains within the 62.1-62.5\% range. This indicates that top-performing models successfully abstract underlying physical principles, treating cultural entities as irrelevant variables regardless of the region.

However, a clear performance drop emerges for mid-tier models. While Gemma3 4B remains relatively stable ($\approx$ 44-45\% across all variants), other models like Phi4 Mini and Qwen3 0.6B experience a drop when moving from the baseline to any cultural variant. Phi4 Mini drops from 30.4\% (baseline) to approximately 23\% across contextual variants, and Qwen3 0.6B drops from 38.6\% to approximately 30-31\%.

Crucially, despite this initial drop, performance across the three contextual variants (Asia, Africa, OcSA) is nearly identical. For example, Phi4 Mini's scores are tightly clustered at 23.6\%, 22.6\%, and 23.1\%, and Qwen3 0.6B varies by approximately 1\% across regions (30.7\%, 30.1\%, 31.1\%). This demonstrates that while some models struggle with the presence of added context, the specific cultural framing, whether Asian, African, or South American, does not differentially impact reasoning performance.

\subsection{Error Analysis}

\begin{figure}[h]
    \centering
    \includegraphics[width=0.47\textwidth]{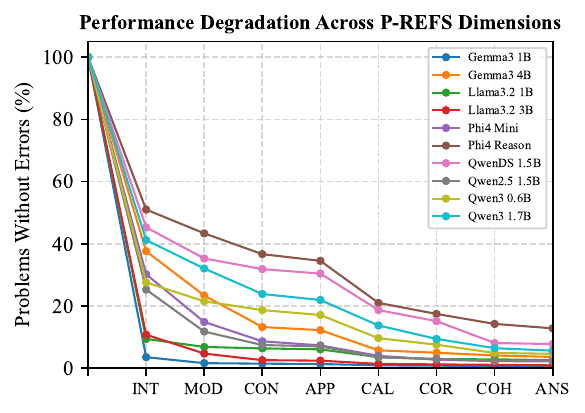}
    \caption{Progressive degradation across P-REFS error categories for problem-solving questions. The y-axis shows the percentage of questions remaining error-free.}
    \label{fig:error_survival}
\end{figure}

Figure~\ref{fig:error_survival} presents a progressive degradation analysis, where each point represents the percentage of problem-solving questions that remain error-free as evaluation proceeds sequentially through the P-REFS error categories.

\paragraph{Early stage errors:} The steepest drops occur at the interpretation (INT) and modeling (MOD) stages, indicating that many failures arise before multi-step computation begins. After INT, weaker models (e.g., Llama3.2 1B, Gemma3 1B, Llama3.2 3B) retain fewer than 10\% error-free solutions, while mid-tier models (e.g., Qwen3 0.6B, Qwen2.5 1.5B) drop to roughly 25-30\%. MOD further reduces the remaining pool, reflecting frequent mismatches between problem understanding and the chosen strategy or assumptions. In contrast, stronger models retain substantially higher error-free fractions after formulation, indicating more reliable setup and strategy selection.

\paragraph{Execution errors:} After the conceptual checks (CON, APP), weaker models remain near zero, while stronger models still retain a non-trivial error-free fraction but lose most remaining solutions at calculation (CAL) and correctness (COR). This pattern suggests that the dominant bottleneck shifts with model strength: weaker models fail primarily at early formulation, whereas stronger models are limited by procedural reliability in multi-step execution. Coherence (COH) contributes comparatively little additional loss once earlier stages are correct. Detailed error frequency analysis is provided in Appendix~\ref{appen_error_prevalence}.

\section{Discussion}

Our results show that final-answer accuracy substantially overestimates reasoning validity on problem-solving questions: even the strongest model (Phi4~Reason) attains only $\approx$13-15\% fully correct solutions despite much higher answer accuracy, and most final-answer-correct outputs still contain at least one reasoning error. This discrepancy persists across model families and sizes, indicating that scaling alone does not yield reliable multi-step quantitative reasoning.

Failure modes also shift with capability. Weaker models most often break during interpretation and modeling, whereas stronger models more frequently fail during downstream execution (notably calculation). Reasoning-oriented training reduces early-stage formulation failures more than execution-stage errors, motivating hybrid approaches that add procedural verification (e.g., arithmetic and unit checks) and stage-targeted interventions informed by P-REFS such as selective-hinting or distillation-based guidance (e.g., Mentor-KD \cite{lee2024mentor}) and training paradigms that explicitly integrate reasoning objectives (e.g., RARE \cite{tran2025rare}).

Across Bloom’s Taxonomy, performance degrades with increasing cognitive demand and is weakest on procedural knowledge. In contrast, top models are largely stable under culturally contextualized variants when underlying physics is preserved, suggesting that contextual rewrites primarily affect mid-tier models. Overall, these findings highlight the need for process-aware monitoring in educational deployments rather than relying on final answers alone.

\section{Conclusion}

This work presents a multi-dimensional evaluation of physics reasoning in contemporary small language models, spanning model families, physics topics, Bloom’s Taxonomy dimensions, and culturally contextualized variants. Across problem-solving questions, we find a consistent gap between answer correctness and reasoning validity: even the strongest model achieves only $\approx$13-15\% fully correct end-to-end solutions, and among correctly answered questions, roughly $\approx$75-98\% contain at least one reasoning error in formulation or execution. Failure modes vary by model strength: weaker models break down predominantly during problem interpretation and modeling, whereas stronger models more often fail during procedural execution, particularly calculation and correctness.

These findings have implications for both model development and educational deployment. On the development side, optimizing for final-answer correctness alone is insufficient; progress requires improving procedural robustness and end-to-end reasoning validity. Promising directions include reasoning-oriented training (effective for formulation), explicit verification mechanisms (for arithmetic and units), and hybrid approaches that integrate symbolic constraints. For educational use, SLMs may offer accessibility and contextual adaptability, but their tendency to produce correct final answers with flawed reasoning risks reinforcing superficial learning unless reasoning quality is explicitly checked. In tutoring contexts, formulation accuracy is especially critical for guiding students through problem setup, and high formulation-error rates in weaker models limit their suitability without additional safeguards.

Bridging these gaps could move SLMs from producing plausible outputs to providing reliable reasoning support, enabling broader access to physics instruction while preserving pedagogical integrity.

\section*{Limitations}

The LLM-as-a-judge framework enables scalable evaluation, but it may introduce 
systematic bias in scoring across P-REFS criteria and in cases requiring judgment 
on numerical precision, unit conversions, and algebraic equivalence. We used expert validation to select the judge with the highest alignment among candidates; however, this validation was conducted on a limited subset and may not fully reflect judge behavior across the full benchmark, all model families, and all rubric categories. Future work should assess cross-judge robustness and incorporate hybrid, rule-based verifiers (e.g., arithmetic and unit normalization checks) to triangulate rubric scores.

Additionally, our cultural contextualization does not capture diversity within countries. While we employed self-consistency checks during contextualization, future work can expand coverage to additional regions and languages, model sub-national variation, and implement more comprehensive automated verification of formula and parameter preservation to better characterize robustness to culturally grounded framing.

\section*{Ethics Statement}

We obtained necessary permissions from OpenStax for the use of their high school physics textbook content in our evaluation of SLMs. \textit{PhysBench} is derived from exercises in OpenStax textbooks: High School Physics \cite{openstax} and Physics for AP{\textregistered} Courses \cite{ap_physics}, both licensed under Creative Commons Attribution (CC BY 4.0). Our modifications include restructured solutions, Bloom's Taxonomy annotations, and culturally contextualized variants. The dataset creation and model evaluation processes were designed to respect intellectual property rights while facilitating research on physics reasoning capabilities. We will release PhysBench under CC BY 4.0, with the restriction that it may not be used 
for training large language models or ingested into generative AI offerings without OpenStax's explicit permission, in accordance with their usage policy. All accompanying code and evaluation scripts will be released under the MIT license to facilitate reproducibility and minimize redundant computational costs.

\bibliography{acl_latex}

\begin{thebibliography}{52}
\providecommand{\natexlab}[1]{#1}

\bibitem[{Abdin et~al.(2025)Abdin, Agarwal, Awadallah, Balachandran, Behl, Chen, de~Rosa, Gunasekar, Javaheripi, Joshi et~al.}]{phi4reasoning}
Marah Abdin, Sahaj Agarwal, Ahmed Awadallah, Vidhisha Balachandran, Harkirat Behl, Lingjiao Chen, Gustavo de~Rosa, Suriya Gunasekar, Mojan Javaheripi, Neel Joshi, and 1 others. 2025.
\newblock Phi-4-reasoning technical report.
\newblock \emph{arXiv preprint arXiv:2504.21318}.

\bibitem[{Abdin et~al.(2024{\natexlab{a}})Abdin, Aneja, Behl, Bubeck, Eldan, Gunasekar, Harrison, Hewett, Javaheripi, Kauffmann et~al.}]{phi4}
Marah Abdin, Jyoti Aneja, Harkirat Behl, S{\'e}bastien Bubeck, Ronen Eldan, Suriya Gunasekar, Michael Harrison, Russell~J Hewett, Mojan Javaheripi, Piero Kauffmann, and 1 others. 2024{\natexlab{a}}.
\newblock Phi-4 technical report.
\newblock \emph{arXiv preprint arXiv:2412.08905}.

\bibitem[{Abdin et~al.(2024{\natexlab{b}})Abdin, Jacobs, Awan, Aneja, Awadallah, Awadalla, Bach, Bahree, Bakhtiari, Behl et~al.}]{phi3}
Marah Abdin, Sam~Ade Jacobs, Ammar~Ahmad Awan, Jyoti Aneja, Ahmed Awadallah, Hany Awadalla, Nguyen Bach, Amit Bahree, Arash Bakhtiari, Harkirat Behl, and 1 others. 2024{\natexlab{b}}.
\newblock Phi-3 technical report: A highly capable language model locally on your phone.
\newblock \emph{arXiv preprint arXiv:2404.14219}.

\bibitem[{Anand et~al.(2023)Anand, Goel, Hira, Buldeo, Kumar, Verma, Gupta, and Shah}]{anand2023sciphyrag}
Avinash Anand, Arnav Goel, Medha Hira, Snehal Buldeo, Jatin Kumar, Astha Verma, Rushali Gupta, and Rajiv~Ratn Shah. 2023.
\newblock Sciphyrag-retrieval augmentation to improve llms on physics q \&a.
\newblock In \emph{International Conference on Big Data Analytics}, pages 50--63. Springer.

\bibitem[{Arora et~al.(2023)Arora, Singh et~al.}]{arora2023have}
Daman Arora, Himanshu~Gaurav Singh, and 1 others. 2023.
\newblock Have llms advanced enough? a challenging problem solving benchmark for large language models.
\newblock \emph{arXiv preprint arXiv:2305.15074}.

\bibitem[{Byrt et~al.(1993)Byrt, Bishop, and Carlin}]{byrt1993bias}
Ted Byrt, Janet Bishop, and John~B Carlin. 1993.
\newblock Bias, prevalence and kappa.
\newblock \emph{Journal of clinical epidemiology}, 46(5):423--429.

\bibitem[{Chang et~al.(2024)Chang, Wang, Wang, Wu, Yang, Zhu, Chen, Yi, Wang, Wang et~al.}]{llmsurvey}
Yupeng Chang, Xu~Wang, Jindong Wang, Yuan Wu, Linyi Yang, Kaijie Zhu, Hao Chen, Xiaoyuan Yi, Cunxiang Wang, Yidong Wang, and 1 others. 2024.
\newblock A survey on evaluation of large language models.
\newblock \emph{ACM transactions on intelligent systems and technology}, 15(3):1--45.

\bibitem[{Chiang and Lee(2023)}]{chiang2023can}
Cheng-Han Chiang and Hung-Yi Lee. 2023.
\newblock Can large language models be an alternative to human evaluations?
\newblock In \emph{Proceedings of the 61st Annual Meeting of the Association for Computational Linguistics (Volume 1: Long Papers)}, pages 15607--15631.

\bibitem[{Cordova and Lepper(1996)}]{cordova1996intrinsic}
Diana~I Cordova and Mark~R Lepper. 1996.
\newblock Intrinsic motivation and the process of learning: Beneficial effects of contextualization, personalization, and choice.
\newblock \emph{Journal of educational psychology}, 88(4):715.

\bibitem[{Fang et~al.(2025)Fang, Wan, Lu, Xing, and Zou}]{fang2025mathodyssey}
Meng Fang, Xiangpeng Wan, Fei Lu, Fei Xing, and Kai Zou. 2025.
\newblock Mathodyssey: Benchmarking mathematical problem-solving skills in large language models using odyssey math data.
\newblock \emph{Scientific Data}, 12(1):1392.

\bibitem[{Feinstein and Cicchetti(1990)}]{feinstein1990high}
Alvan~R Feinstein and Domenic~V Cicchetti. 1990.
\newblock High agreement but low kappa: I. the problems of two paradoxes.
\newblock \emph{Journal of clinical epidemiology}, 43(6):543--549.

\bibitem[{Grattafiori et~al.(2024)Grattafiori, Dubey, Jauhri, Pandey, Kadian, Al-Dahle, Letman, Mathur, Schelten, Vaughan et~al.}]{llama3}
Aaron Grattafiori, Abhimanyu Dubey, Abhinav Jauhri, Abhinav Pandey, Abhishek Kadian, Ahmad Al-Dahle, Aiesha Letman, Akhil Mathur, Alan Schelten, Alex Vaughan, and 1 others. 2024.
\newblock The llama 3 herd of models.
\newblock \emph{arXiv preprint arXiv:2407.21783}.

\bibitem[{Gu et~al.(2024)Gu, Jiang, Shi, Tan, Zhai, Xu, Li, Shen, Ma, Liu et~al.}]{judgesurvey}
Jiawei Gu, Xuhui Jiang, Zhichao Shi, Hexiang Tan, Xuehao Zhai, Chengjin Xu, Wei Li, Yinghan Shen, Shengjie Ma, Honghao Liu, and 1 others. 2024.
\newblock A survey on llm-as-a-judge.
\newblock \emph{arXiv preprint arXiv:2411.15594}.

\bibitem[{Guo et~al.(2025)Guo, Yang, Zhang, Song, Zhang, Xu, Zhu, Ma, Wang, Bi et~al.}]{deepseek}
Daya Guo, Dejian Yang, Haowei Zhang, Junxiao Song, Ruoyu Zhang, Runxin Xu, Qihao Zhu, Shirong Ma, Peiyi Wang, Xiao Bi, and 1 others. 2025.
\newblock Deepseek-r1: Incentivizing reasoning capability in llms via reinforcement learning.
\newblock \emph{arXiv preprint arXiv:2501.12948}.

\bibitem[{Gwet(2008)}]{gwet2008computing}
Kilem~Li Gwet. 2008.
\newblock Computing inter-rater reliability and its variance in the presence of high agreement.
\newblock \emph{British Journal of Mathematical and Statistical Psychology}, 61(1):29--48.

\bibitem[{He et~al.(2024)He, Luo, Bai, Hu, Thai, Shen, Hu, Han, Huang, Zhang et~al.}]{he2024olympiadbench}
Chaoqun He, Renjie Luo, Yuzhuo Bai, Shengding Hu, Zhen~Leng Thai, Junhao Shen, Jinyi Hu, Xu~Han, Yujie Huang, Yuxiang Zhang, and 1 others. 2024.
\newblock Olympiadbench: A challenging benchmark for promoting agi with olympiad-level bilingual multimodal scientific problems.
\newblock \emph{arXiv preprint arXiv:2402.14008}.

\bibitem[{Hendrycks et~al.(2020)Hendrycks, Burns, Basart, Zou, Mazeika, Song, and Steinhardt}]{hendrycks2020measuring}
Dan Hendrycks, Collin Burns, Steven Basart, Andy Zou, Mantas Mazeika, Dawn Song, and Jacob Steinhardt. 2020.
\newblock Measuring massive multitask language understanding.
\newblock \emph{arXiv preprint arXiv:2009.03300}.

\bibitem[{Huang et~al.(2023)Huang, Bai, Zhu, Zhang, Zhang, Su, Liu, Lv, Zhang, Fu et~al.}]{huang2023c}
Yuzhen Huang, Yuzhuo Bai, Zhihao Zhu, Junlei Zhang, Jinghan Zhang, Tangjun Su, Junteng Liu, Chuancheng Lv, Yikai Zhang, Yao Fu, and 1 others. 2023.
\newblock C-eval: A multi-level multi-discipline chinese evaluation suite for foundation models.
\newblock \emph{Advances in Neural Information Processing Systems}, 36:62991--63010.

\bibitem[{Jiang et~al.(2025)Jiang, Chen, Cao, Lee, and Tan}]{codejudgebench2025}
Hongchao Jiang, Yiming Chen, Yushi Cao, Hung-yi Lee, and Robby~T Tan. 2025.
\newblock Codejudgebench: Benchmarking llm-as-a-judge for coding tasks.
\newblock \emph{arXiv preprint arXiv:2507.10535}.

\bibitem[{Karim et~al.(2025)Karim, Karim, Lohana, Keon, Singh, and Sattar}]{karim2025lost}
Aabid Karim, Abdul Karim, Bhoomika Lohana, Matt Keon, Jaswinder Singh, and Abdul Sattar. 2025.
\newblock Lost in cultural translation: Do llms struggle with math across cultural contexts?
\newblock \emph{arXiv preprint arXiv:2503.18018}.

\bibitem[{Krathwohl(2002)}]{krathwohl2002revision}
David~R Krathwohl. 2002.
\newblock A revision of bloom's taxonomy: An overview.
\newblock \emph{Theory into practice}, 41(4):212--218.

\bibitem[{Lanham et~al.(2023)Lanham, Chen, Radhakrishnan, Steiner, Denison, Hernandez, Li, Durmus, Hubinger, Kernion et~al.}]{lanham2023measuring}
Tamera Lanham, Anna Chen, Ansh Radhakrishnan, Benoit Steiner, Carson Denison, Danny Hernandez, Dustin Li, Esin Durmus, Evan Hubinger, Jackson Kernion, and 1 others. 2023.
\newblock Measuring faithfulness in chain-of-thought reasoning.
\newblock \emph{arXiv preprint arXiv:2307.13702}.

\bibitem[{Lee et~al.(2024)Lee, Kim, and Lee}]{lee2024mentor}
Hojae Lee, Junho Kim, and SangKeun Lee. 2024.
\newblock Mentor-{KD}: Making small language models better multi-step reasoners.
\newblock In \emph{Proceedings of the 2024 Conference on Empirical Methods in Natural Language Processing}, pages 17643--17658, Miami, Florida, USA. Association for Computational Linguistics.

\bibitem[{Lee and Hockenmaier(2025)}]{lee2025evaluating}
Jinu Lee and Julia Hockenmaier. 2025.
\newblock Evaluating step-by-step reasoning traces: A survey.
\newblock \emph{arXiv preprint arXiv:2502.12289}.

\bibitem[{Leinonen et~al.(2021)Leinonen, Denny, and Whalley}]{leinonen2021exploring}
Juho Leinonen, Paul Denny, and Jacqueline Whalley. 2021.
\newblock Exploring the effects of contextualized problem descriptions on problem solving.
\newblock In \emph{Proceedings of the 23rd Australasian Computing Education Conference}, pages 30--39.

\bibitem[{Lu et~al.(2022)Lu, Mishra, Xia, Qiu, Chang, Zhu, Tafjord, Clark, and Kalyan}]{lu2022learn}
Pan Lu, Swaroop Mishra, Tanglin Xia, Liang Qiu, Kai-Wei Chang, Song-Chun Zhu, Oyvind Tafjord, Peter Clark, and Ashwin Kalyan. 2022.
\newblock Learn to explain: Multimodal reasoning via thought chains for science question answering.
\newblock \emph{Advances in Neural Information Processing Systems}, 35:2507--2521.

\bibitem[{Marchee and Joje~Mar(2022)}]{marchee2022effectiveness}
T~Marchee and P~Joje~Mar. 2022.
\newblock Effectiveness of contextualization in science instruction to enhance science literacy in the philippines: A meta-analysis.
\newblock \emph{International Journal of Learning, Teaching and Educational Research}, 21(1):140--156.

\bibitem[{Mirzadeh et~al.(2024)Mirzadeh, Alizadeh, Shahrokhi, Tuzel, Bengio, and Farajtabar}]{mirzadeh2024gsm}
Iman Mirzadeh, Keivan Alizadeh, Hooman Shahrokhi, Oncel Tuzel, Samy Bengio, and Mehrdad Farajtabar. 2024.
\newblock Gsm-symbolic: Understanding the limitations of mathematical reasoning in large language models.
\newblock \emph{arXiv preprint arXiv:2410.05229}.

\bibitem[{Mushtaq et~al.(2025)Mushtaq, Taj, Naeem, Ghaznavi, and Qadir}]{mushtaq2025worldview}
Abdullah Mushtaq, Imran Taj, Rafay Naeem, Ibrahim Ghaznavi, and Junaid Qadir. 2025.
\newblock Worldview-bench: A benchmark for evaluating global cultural perspectives in large language models.
\newblock \emph{arXiv preprint arXiv:2505.09595}.

\bibitem[{Qwen et~al.(2025)Qwen, :, Yang, Yang, Zhang, Hui, Zheng, Yu, Li, Liu, Huang, Wei, Lin, Yang, Tu, Zhang, Yang, Yang, Zhou, Lin, Dang, Lu, Bao, Yang, Yu, Li, Xue, Zhang, Zhu, Men, Lin, Li, Tang, Xia, Ren, Ren, Fan, Su, Zhang, Wan, Liu, Cui, Zhang, and Qiu}]{qwen2_5}
Qwen, :, An~Yang, Baosong Yang, Beichen Zhang, Binyuan Hui, Bo~Zheng, Bowen Yu, Chengyuan Li, Dayiheng Liu, Fei Huang, Haoran Wei, Huan Lin, Jian Yang, Jianhong Tu, Jianwei Zhang, Jianxin Yang, Jiaxi Yang, Jingren Zhou, and 25 others. 2025.
\newblock Qwen2.5 technical report.
\newblock \emph{arXiv preprint arXiv:2412.15115}.

\bibitem[{Ramezani and Xu(2023)}]{ramezani2023knowledge}
Aida Ramezani and Yang Xu. 2023.
\newblock Knowledge of cultural moral norms in large language models.
\newblock \emph{arXiv preprint arXiv:2306.01857}.

\bibitem[{Samo et~al.(2018)Samo, Kartasasmita et~al.}]{samo2018culture}
Damianus~Dao Samo, Bana~G Kartasasmita, and 1 others. 2018.
\newblock Culture-based contextual learning to increase problem-solving ability of first year university student.
\newblock \emph{Journal on Mathematics Education}, 9(1):81--94.

\bibitem[{Shojaee et~al.(2025)Shojaee, Mirzadeh, Alizadeh, Horton, Bengio, and Farajtabar}]{shojaee2025illusion}
Parshin Shojaee, Iman Mirzadeh, Keivan Alizadeh, Maxwell Horton, Samy Bengio, and Mehrdad Farajtabar. 2025.
\newblock The illusion of thinking: Understanding the strengths and limitations of reasoning models via the lens of problem complexity.
\newblock \emph{arXiv preprint arXiv:2506.06941}.

\bibitem[{Siddique et~al.(2025)Siddique, Alam, Rafy, Raiyan, Mahmud, and Hasan}]{physicseval}
Oshayer Siddique, JM~Alam, Md~Jobayer~Rahman Rafy, Syed~Rifat Raiyan, Hasan Mahmud, and Md~Kamrul Hasan. 2025.
\newblock Physicseval: Inference-time techniques to improve the reasoning proficiency of large language models on physics problems.
\newblock \emph{arXiv preprint arXiv:2508.00079}.

\bibitem[{Srivastava et~al.(2025)Srivastava, Cao, and Wang}]{srivastava2025towards}
Gaurav Srivastava, Shuxiang Cao, and Xuan Wang. 2025.
\newblock Towards reasoning ability of small language models.
\newblock \emph{arXiv preprint arXiv:2502.11569}.

\bibitem[{Sun et~al.(2020)Sun, Yu, Song, Liu, Yang, and Zhou}]{sun2020mobilebert}
Zhiqing Sun, Hongkun Yu, Xiaodan Song, Renjie Liu, Yiming Yang, and Denny Zhou. 2020.
\newblock Mobilebert: a compact task-agnostic bert for resource-limited devices.
\newblock In \emph{Proceedings of the 58th Annual Meeting of the Association for Computational Linguistics}, pages 2158--2170.

\bibitem[{Tan et~al.(2025)Tan, Zhuang, Montgomery, Tang, Cuadron, Wang, Popa, and Stoica}]{judgebench}
Sijun Tan, Siyuan Zhuang, Kyle Montgomery, William~Yuan Tang, Alejandro Cuadron, Chenguang Wang, Raluca Popa, and Ion Stoica. 2025.
\newblock Judgebench: A benchmark for evaluating {LLM}-based judges.
\newblock In \emph{The Thirteenth International Conference on Learning Representations}.

\bibitem[{Team et~al.(2023)Team, Anil, Borgeaud, Alayrac, Yu, Soricut, Schalkwyk, Dai, Hauth, Millican et~al.}]{team2023gemini}
Gemini Team, Rohan Anil, Sebastian Borgeaud, Jean-Baptiste Alayrac, Jiahui Yu, Radu Soricut, Johan Schalkwyk, Andrew~M Dai, Anja Hauth, Katie Millican, and 1 others. 2023.
\newblock Gemini: a family of highly capable multimodal models.
\newblock \emph{arXiv preprint arXiv:2312.11805}.

\bibitem[{Team et~al.(2025)Team, Kamath, Ferret, Pathak, Vieillard, Merhej, Perrin, Matejovicova, Ram{\'e}, Rivi{\`e}re et~al.}]{gemma3}
Gemma Team, Aishwarya Kamath, Johan Ferret, Shreya Pathak, Nino Vieillard, Ramona Merhej, Sarah Perrin, Tatiana Matejovicova, Alexandre Ram{\'e}, Morgane Rivi{\`e}re, and 1 others. 2025.
\newblock Gemma 3 technical report.
\newblock \emph{arXiv preprint arXiv:2503.19786}.

\bibitem[{Thakur et~al.(2024)Thakur, Choudhary, Ramayapally, Vaidyanathan, and Hupkes}]{thakur2024judging}
Aman~Singh Thakur, Kartik Choudhary, Venkat~Srinik Ramayapally, Sankaran Vaidyanathan, and Dieuwke Hupkes. 2024.
\newblock Judging the judges: Evaluating alignment and vulnerabilities in llms-as-judges.
\newblock \emph{arXiv preprint arXiv:2406.12624}.

\bibitem[{Tomar et~al.(2025)Tomar, Sahoo, Mittal, Murthy, and Bhattacharyya}]{tomar2025mathematics}
Aditya Tomar, Nihar~Ranjan Sahoo, Ashish Mittal, Rudra Murthy, and Pushpak Bhattacharyya. 2025.
\newblock Mathematics isn't culture-free: Probing cultural gaps via entity and scenario perturbations.
\newblock \emph{arXiv preprint arXiv:2507.00883}.

\bibitem[{Tran et~al.(2025)Tran, Yao, Yang, Wang, Zhang, Han, Ouyang, and Yu}]{tran2025rare}
Hieu Tran, Zonghai Yao, Zhichao Yang, Junda Wang, Yifan Zhang, Shuo Han, Feiyun Ouyang, and Hong Yu. 2025.
\newblock {RARE}: Retrieval-augmented reasoning enhancement for large language models.
\newblock In \emph{Proceedings of the 63rd Annual Meeting of the Association for Computational Linguistics (Volume 1: Long Papers)}, pages 18305--18330, Vienna, Austria. Association for Computational Linguistics.

\bibitem[{Urone and Hinrichs(2020)}]{openstax}
Paul~Peter Urone and Roger Hinrichs. 2020.
\newblock \href {https://openstax.org/books/physics} {\emph{Physics}}.
\newblock OpenStax, Houston, Texas.

\bibitem[{Veselovsky et~al.(2025)Veselovsky, Argin, Stroebl, Wendler, West, Evans, Griffiths, and Narayanan}]{veselovsky2025localized}
Veniamin Veselovsky, Berke Argin, Benedikt Stroebl, Chris Wendler, Robert West, James Evans, Thomas~L Griffiths, and Arvind Narayanan. 2025.
\newblock Localized cultural knowledge is conserved and controllable in large language models.
\newblock \emph{arXiv preprint arXiv:2504.10191}.

\bibitem[{Wang et~al.(2025)Wang, Su, Liu, Li, Xia, Xiao, Zhang, Dai, Chen, Meng et~al.}]{wang2025physunibench}
Lintao Wang, Encheng Su, Jiaqi Liu, Pengze Li, Peng Xia, Jiabei Xiao, Wenlong Zhang, Xinnan Dai, Xi~Chen, Yuan Meng, and 1 others. 2025.
\newblock Physunibench: An undergraduate-level physics reasoning benchmark for multimodal models.
\newblock \emph{arXiv preprint arXiv:2506.17667}.

\bibitem[{Wang et~al.(2023)Wang, Hu, Lu, Zhu, Zhang, Subramaniam, Loomba, Zhang, Sun, and Wang}]{wang2023scibench}
Xiaoxuan Wang, Ziniu Hu, Pan Lu, Yanqiao Zhu, Jieyu Zhang, Satyen Subramaniam, Arjun~R Loomba, Shichang Zhang, Yizhou Sun, and Wei Wang. 2023.
\newblock Scibench: Evaluating college-level scientific problem-solving abilities of large language models.
\newblock \emph{arXiv preprint arXiv:2307.10635}.

\bibitem[{Wei et~al.(2025)Wei, Carvalho, and Stamper}]{wei2025small}
Yumou Wei, Paulo Carvalho, and John Stamper. 2025.
\newblock Small but significant: On the promise of small language models for accessible aied.
\newblock \emph{arXiv preprint arXiv:2505.08588}.

\bibitem[{Whitehouse et~al.(2025)Whitehouse, Wang, Yu, Li, Weston, Kulikov, and Saha}]{j1reinforcement2025}
Chenxi Whitehouse, Tianlu Wang, Ping Yu, Xian Li, Jason Weston, Ilia Kulikov, and Swarnadeep Saha. 2025.
\newblock J1: Incentivizing thinking in llm-as-a-judge via reinforcement learning.
\newblock \emph{arXiv preprint arXiv:2505.10320}.

\bibitem[{Wolfe et~al.(2022)Wolfe, Gasper, Stoke, Kretchman, Anderson, Czuba, Oberoi, Pujji, Lyublinskaya, and Ingram}]{ap_physics}
Gregg Wolfe, Erika Gasper, John Stoke, Julie Kretchman, David Anderson, Nathan Czuba, Sudhi Oberoi, Liza Pujji, Irina Lyublinskaya, and Douglas Ingram. 2022.
\newblock \href {https://openstax.org/books/college-physics-ap-courses-2e/pages/1-connection-for-ap-r-courses} {\emph{College Physics for AP{\textregistered} Courses}}, 2 edition.
\newblock OpenStax, Houston, Texas.
\newblock Available at: \url{https://openstax.org/books/college-physics-ap-courses-2e}.

\bibitem[{Xia et~al.(2025)Xia, Li, Liu, Wu, and Liu}]{xia2025evaluating}
Shijie Xia, Xuefeng Li, Yixin Liu, Tongshuang Wu, and Pengfei Liu. 2025.
\newblock Evaluating mathematical reasoning beyond accuracy.
\newblock In \emph{Proceedings of the AAAI Conference on Artificial Intelligence}, volume~39, pages 27723--27730.

\bibitem[{Yang et~al.(2025)Yang, Li, Yang, Zhang, Hui, Zheng, Yu, Gao, Huang, Lv et~al.}]{qwen3}
An~Yang, Anfeng Li, Baosong Yang, Beichen Zhang, Binyuan Hui, Bo~Zheng, Bowen Yu, Chang Gao, Chengen Huang, Chenxu Lv, and 1 others. 2025.
\newblock Qwen3 technical report.
\newblock \emph{arXiv preprint arXiv:2505.09388}.

\bibitem[{Zhang et~al.(2025)Zhang, Dong, Wu, Huang, Jia, Fernando, Shou, Zhang, and Liu}]{zhang2025physreason}
Xinyu Zhang, Yuxuan Dong, Yanrui Wu, Jiaxing Huang, Chengyou Jia, Basura Fernando, Mike~Zheng Shou, Lingling Zhang, and Jun Liu. 2025.
\newblock Physreason: A comprehensive benchmark towards physics-based reasoning.
\newblock \emph{arXiv preprint arXiv:2502.12054}.

\end{thebibliography}

\clearpage
\appendix
\onecolumn

\startcontents[appendixtoc]
\printcontents[appendixtoc]{l}{1}{\section*{Appendix}}

\twocolumn

\section{Dataset Development and Annotation}

\subsection{Dataset extraction and preprocessing} \label{data_appendix}

A challenge encountered during the data extraction phase was the representation of mathematical equations, which were predominantly embedded as images rather than text within the exercise documents. To address this challenge, we used multiple tools, including specialized optical character recognition (OCR) models (Google Vision API, Mathpix), to convert equation images into \LaTeX{} format. This approach significantly reduced the need for manual transcription of mathematical notation and preserved the integrity of the mathematical content. 

Questions that contained essential images were handled systematically: those where the visual content could not be adequately represented in text were removed from the dataset, while questions where images could be described textually were retained with detailed descriptions of the visual elements incorporated into the question text. Additionally, for problems originally requiring graphical outputs (e.g., plotting a trajectory), the questions were modified wherever possible to request text-based descriptions of the graph's key features (such as slope or intercept); questions where such textual conversion was not feasible were excluded from the dataset.

\paragraph{Answer restructuring:} The raw dataset comprising questions, answer choices, correct answers, and reasoning was sourced from OpenStax textbook and teacher resources. To facilitate a granular evaluation of reasoning, we utilized Google's Gemini 2.5 Pro to restructure these raw solutions into a standardized step-by-step format. This automated process involved parsing the original narrative explanations and segmenting them into discrete, logical units.
This transformation ensured that every solution adhered to a consistent template, making the reasoning process explicit and easier to evaluate. This format was designed with the idea that each step contains a complete unit of logical reasoning with all necessary context, can be understood and evaluated independently, and provides meaningful advancement toward the final solution. The structured answer format is illustrated in Figure~\ref{structured_format}.

\begin{figure}[h]
\centering
\begin{lstlisting}[
    style=mystyle,
    morekeywords={}
]
Problem Formulation:

Given: <What is given in the question>
Find: <What needs to be found or solved for in the question>
Approach: <What approach is being taken to solve the problem>
Assumptions: <What are things you are assuming to solve the problem>

Step-by-step Detailed Solution:

Step 1: <What's the step>
Reasoning: <Why are you following this step>
General Formula: <What's the formula to solve this problem>
Substitution: <Substitute the given values in the problem to solve it>
Calculation: <Do the calculations here and solve to get the final answer>
Result: <Final Answer from the step>

Step 2: <What's the step>
Reasoning: <Why are you following this step>
General Formula: <What's the formula to solve this problem>
Substitution: <Substitute the given values in the problem to solve it>
Calculation: <Do the calculations here and solve to get the final answer>
Result: <Final Answer from the step>

[Continue with additional steps as needed]

Final Answer: <Final answer to the entire question>
\end{lstlisting}
\caption{Structured answer template used for reformatting textbook solutions.}
\label{structured_format}
\end{figure}

This structured format facilitates fine-grained analysis of model reasoning at each stage of problem-solving, enabling identification of where errors occur in the reasoning chain.

\paragraph{Human verification and quality control:} Following the automated restructuring, the entire dataset underwent a manual verification process conducted by graduate and undergraduate engineering students. This expert-led review was critical for identifying and rectifying inconsistencies present in the source material, such as references to non-existent figures, logical flaws in the provided explanations, or incomplete problem statements. The final reference solutions were human-verified; when necessary, verifiers corrected equations, assumptions, intermediate steps, and units to ensure physical and mathematical validity. Entries with irreparable flaws (e.g., missing essential information such that the intended problem could not be unambiguously reconstructed) were removed, while the remaining items were revised to meet the quality standards required for this study.

\subsection{Bloom's taxonomy annotations} \label{blooms_appendix}
We annotated each physics problem according to the revised Bloom's Taxonomy \cite{krathwohl2002revision}, which consists of two dimensions: \\

\noindent \textbf{Cognitive Process Dimension:} 

\begin{itemize} 
\item \textit{Remember:} Retrieving relevant knowledge from long-term memory, such as recalling facts, basic concepts, or definitions 
\item \textit{Understand:} Constructing meaning from instructional materials, including interpreting, summarizing, and explaining ideas 
\item \textit{Apply:} Using procedures or learned methods in a given situation to solve problems or carry out tasks 
\item \textit{Analyze:} Breaking down information into components, identifying relationships or patterns, and understanding structure and function 
\item \textit{Evaluate:} Judging or determining the value of material or methods based on criteria or standards 
\item \textit{Create:} Generating new ideas, products, or structures by combining elements into a coherent or functional whole 
\end{itemize}
\textbf{Knowledge Dimension:} 
\begin{itemize} \item \textit{Factual:} The basic elements or facts students must know to solve problems, including terminology and specific details 
\item \textit{Conceptual:} The interrelationships between elements, such as theories, principles, and models, that enable function within a domain 
\item \textit{Procedural:} Knowing how to perform tasks, techniques, and methods, and when to apply them 
\item \textit{Metacognitive:} Knowledge about one's own cognition and how to regulate it, including self-awareness of learning strategies 
\end{itemize}

\paragraph{Annotation methodology:} The classification of questions according to these dimensions was automated using Google's Gemini 2.5 Flash. To validate the reliability of these automated annotations, a stratified random subset of approximately 300 questions, spanning diverse topics, skills, and knowledge dimensions, was manually reviewed by human experts. The human-model inter-annotator agreement was found to be approximately 95\%, indicating high consistency and reliability in the automated classification.

The distribution of questions across the Cognitive Process and Knowledge Dimensions are give in Tables~\ref{cognitive} and \ref{knowledge}.

\begin{table}[h]
\centering
\caption{Distribution of questions across Cognitive Process Dimension}
\label{cognitive}
\begin{tabular}{lcc}
\toprule
\textbf{Cognitive Level} & \textbf{Percentage} & \textbf{Count} \\
\midrule
Remember & 6.8\% & 216 \\
Understand & 13.6\% & 429 \\
Apply & 62.7\% & 1981 \\
Analyze & 14.0\% & 443 \\
Evaluate & 1.6\% & 52 \\
Create & 1.3\% & 41 \\
\bottomrule
\end{tabular}
\end{table}

\begin{table}[h]
\centering
\caption{Distribution of questions across Knowledge Dimension}
\label{knowledge}
\begin{tabular}{lcc}
\toprule
\textbf{Knowledge Type} & \textbf{Percentage} & \textbf{Count} \\
\midrule
Factual & 7.0\% & 222 \\
Conceptual & 23.6\% & 745 \\
Procedural & 69.4\% & 2195 \\
Metacognitive & 0\% & 0 \\
\bottomrule
\end{tabular}
\end{table}

As evidenced by this distribution, the dataset contained questions spanning different levels of cognitive skills. The majority of questions (approximately 63\%) were classified at the Apply level, while foundational skills (Remember, Understand) accounted for roughly 20\%, and complex analytical tasks (Analyze, Evaluate, Create) comprised the remaining 17\%. This distribution reflects the typical emphasis in high school physics education, which prioritizes the application of learned procedures and algorithmic problem-solving.

Notably, no questions were classified as addressing metacognitive knowledge. This absence reflects the traditional focus of high school physics curricula on factual, conceptual, and procedural knowledge rather than on developing students' awareness of their own cognitive processes. Additionally, metacognitive questions are challenging to assess in standardized formats and are often addressed through reflective exercises or learning journals rather than end-of-chapter problems.

\subsection{Topic composition in the dataset}

The distribution of physics topics in our dataset reflects the comprehensive coverage of a typical high school physics curriculum. Table \ref{tab:topic_composition} shows the percentage and count breakdown of questions across different physics domains.

\begin{table}[h]
\centering
\caption{Distribution of questions across physics topics}
\label{tab:topic_composition}
\begin{tabular}{p{3.5cm}cc}
\toprule
\textbf{Physics Topic} & \textbf{Percentage} & \textbf{Count} \\
\midrule
Introduction & 2.1\% & 65 \\
Mechanics & 29.1\% & 919 \\
Electricity \& Magnetism & 20.8\% & 658 \\
Thermodynamics & 9.3\% & 295 \\
Waves \& Acoustics & 8.8\% & 279 \\
Optics & 11.4\% & 361 \\
Modern Physics & 18.5\% & 585 \\
\bottomrule
\end{tabular}
\end{table}

This topic distribution ensured comprehensive coverage of the physics curriculum, allowing us to evaluate SLM performance across the full spectrum of physics concepts typically encountered in high school education. The contextualization process maintained this topic distribution in each culturally adapted dataset, ensuring that comparative analyses across different cultural contexts were not confounded by variations in topic coverage.

The Introduction category includes foundational concepts such as scientific notation, measurement, and dimensional analysis. Mechanics covers motion, forces, energy, and momentum. Electricity \& Magnetism encompasses electric charge, current, circuits, and magnetic fields. Thermodynamics includes heat, temperature, and the laws of thermodynamics. Waves \& Acoustics covers mechanical waves, sound, and basic wave phenomena. Optics includes light, mirrors, lenses, and optical instruments. Modern Physics covers topics such as quantum mechanics, atomic physics, and nuclear physics.

\section{Cultural Contextualization Methodology}\label{context_appendix}

Our cultural contextualization approach required developing comprehensive regional databases to ensure authentic representation. Countries were selected systematically based on the United Nations Geoscheme\footnote{\url{https://unstats.un.org/unsd/methodology/m49/}}, with particular emphasis on underrepresented regions. We organized cultural information into three distinct regional datasets:

\begin{enumerate}
\item \textbf{Asian Context:} Included information from countries such as India, China, Indonesia, Philippines, and many more (51 countries in total)
\item \textbf{African Context:} Incorporated elements from Nigeria, Kenya, South Africa, Ethiopia, and many more (58 countries in total)
\item \textbf{South American and Oceania Context:} Featured Brazil, Argentina, Colombia, Peru, Australia, New Zealand, and many more (41 countries in total)
\end{enumerate}

\subsection{Cultural database generation} 

For each country, we compiled structured information across eight cultural dimensions: names and honorifics, landmarks and geography, traditions and practices, festivals and celebrations, food and cuisines, sports and activities, industries and occupations, and transportation.

We generated the country-level cultural databases using Gemini 2.5 Pro with Google Search grounding enabled (via the Google GenAI API with the \texttt{google\_search} tool). For each country, the model produced structured cultural entries (e.g., landmarks, festivals, foods, sports, transportation) and returned grounding metadata containing retrieved web sources (URLs) associated with the generation. We logged the grounding sources (from \texttt{grounding\_metadata.grounding\_chunks}) for each generation and filtered/flagged outputs that did not return grounding metadata. We additionally performed targeted spot-checks and cross-referencing against widely used public references (e.g., Wikipedia, UNESCO pages, and BBC Country Profiles) to reduce unverifiable content and stereotypical portrayals. The resulting vetted cultural databases were then used as contextual input during the question adaptation phase.

\subsection{Contextual question generation} 

\begin{figure*}[h]
  \centering
  \includegraphics[width=\linewidth]{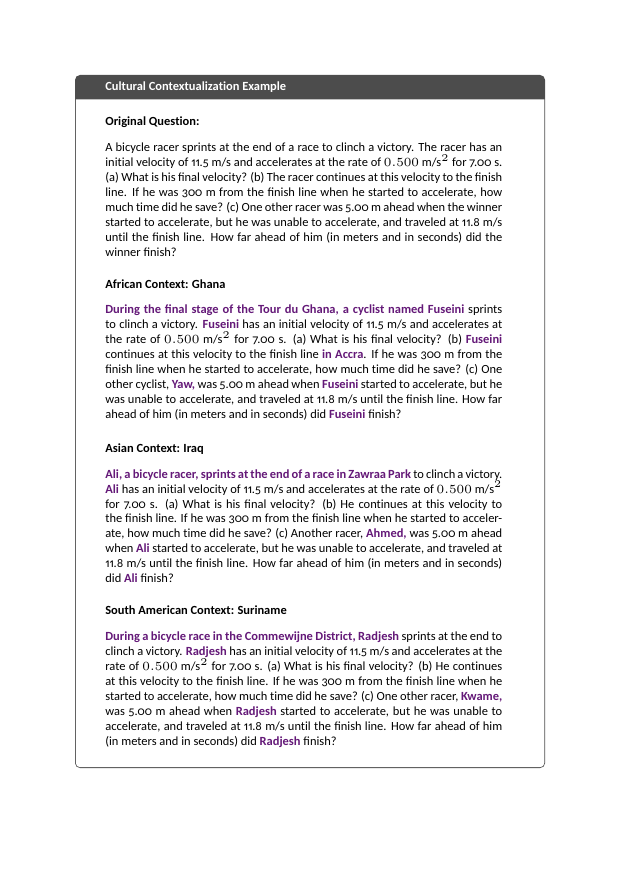}
  \vspace{-3cm}
  \caption{Cultural contextualization example: original and contextual variants (highlighted).}
  \label{fig:cultural_context_example}
\end{figure*}

With these three regional cultural context databases covering 150 countries, we proceeded to the question adaptation phase. While initial pilot experiments were conducted with Gemini 2.5 Flash, we observed that the model occasionally produced physically infeasible scenarios, such as assigning highway velocities to manual rickshaws, subjecting traditional wooden vessels to open flames for boiling, or attributing impossible sustained sprinting speeds to human runners, despite maintaining the correct mathematical structure. To address this, we selected Google's Gemini 2.5 Pro for the final adaptation of the 900 systematically selected questions from \textit{PhysBench}, utilizing its superior reasoning capabilities to maintain both physical validity and cultural authenticity. Figure~\ref{fig:cultural_context_example} illustrates an example of an original question and its culturally contextualized variants, with contextual elements highlighted.

The detailed system prompt used to guide this adaptation process is presented in Figure~\ref{fig:adaptation_prompt}. For each question, the model was instructed to analyze the original scenario, deconstruct its underlying physics principles, and then integrate elements from our cultural context database. A critical directive in this process was to strictly preserve physics fidelity. The model was instructed to ensure that all contextualized variations retained the original core physics concepts, mathematical relationships, formulae (preserved in \LaTeX{} notation), and the overall difficulty level. The system maintained a history of previously generated questions for each country within the region in each generation instance, which helped prevent repetition and ensure authenticity. For multiple choice questions, options and correct answers were generally preserved, unless contextual adaptation required modification for coherence. For open-ended questions, the underlying reasoning remained consistent with the physics tested in the original problem.

We generated five candidate contextualizations per question and checked for consistency across numerical parameters (masses, velocities, forces, distances, etc.), mathematical formulas (preserved in \LaTeX{}), physical principles and constraints, and expected answer values and units. Questions where no consistent candidate could be selected from the five options due to inconsistencies in core physics elements or physically infeasible scenarios were flagged for regeneration. This self-consistency check ensured that cultural rewrites preserved the underlying physics structure while maintaining realistic physical scenarios.

A custom implementation managed the entire workflow from question selection to final output processing. The system maintained comprehensive records of all generation attempts, producing three culturally adapted datasets (\textit{PhysBench}$_{\text{Asia}}$, \textit{PhysBench}$_{\text{Africa}}$, and \textit{PhysBench}$_{\text{OcSA}}$) for comparative analysis of the physics reasoning abilities of SLMs across different cultural contexts.

\newpage

\FloatBarrier
\begin{figure*}[t]
\centering
\begin{lstlisting}[style=promptstyle, basicstyle=\ttfamily\scriptsize]
You are tasked with creating a culturally contextualized physics question.

Input:
    Question: {question}
    Solution: {solution}
    Context: {context}
    Question History: {question_history}

Step-by-step Instructions:

Step 1: Analyze the physics scenario
- Identify the physical objects involved (vehicles, people, containers, etc.)
- Note the physical quantities (speed, mass, distance, time, force, etc.)
- Identify the realistic ranges for these quantities in the original problem
- Understand what type of scenario this is (motion, collision, fluid, thermal, etc.)

Step 2: Categorize replaceable elements
For each element in the original question, identify its CATEGORY and PHYSICAL CONSTRAINTS:
- Person names -> Can replace with any culturally appropriate name
- Vehicles -> Must replace with vehicle of SIMILAR speed/mass capabilities
- Locations -> Must be appropriate for the scenario (highway -> major road, lake -> water body)
- Objects -> Must have similar physical properties (mass, size, material)
- Distances -> Must be realistic for the cultural setting
- Technical/scientific equipment -> Generally should NOT be replaced

Step 3: Review question history
- Check UsedElements to avoid repetition
- Only use elements that have NOT been used before for this country

Step 4: Select culturally appropriate AND physically plausible replacements

CRITICAL RULE: The replacement must be physically capable of the values in the problem.

Example of CORRECT adaptation (Kinematics):
- Original: "A car travels at 90 km/h on a highway for 2 hours."
- For India: "A Maruti Suzuki travels at 90 km/h on the Mumbai-Pune Expressway for 2 hours."
- Why correct: Cars can travel at 90 km/h, expressways allow such speeds.

Example of INCORRECT adaptation:
- Original: "A car travels at 90 km/h on a highway for 2 hours."
- Wrong: "A cycle rickshaw travels at 90 km/h through the streets of Varanasi for 2 hours."
- Why wrong: Cycle rickshaws cannot travel at 90 km/h.

Step 5: Verify physical plausibility BEFORE adapting
For each replacement, ask:
- Can this object/vehicle realistically achieve the values mentioned (speed, mass, temperature, etc.)?
- Would this scenario actually happen in real life?

Step 6: Create the culturally adapted question
- Replace elements ONLY with physically plausible alternatives
- Keep ALL formulas and physics unchanged
- Integrate cultural elements naturally without explanation
- Do NOT add phrases like "traditional", "which is a...", or explanatory text

Step 7: Adapt the complete solution
- Rewrite the Solution with cultural context in the EXACT SAME FORMAT as input
- The Solution must contain these sections in order:
  1. Problem Formulation (update names/locations, keep physics)
  2. Step-by-step Detailed Solution (update context, keep all calculations)
  3. Final Answer (update context, keep numerical answer)

Step 8: Final validation checklist
Before outputting, verify:
- The adapted question makes physical sense
- A reader could solve this without confusion
- All physical values are realistic for the replacements
- No cultural elements are explained or described
- No elements from UsedElements are repeated
- The Solution format matches the input exactly
- All physics and mathematics are preserved

Output format (return as JSON):
{
  "Country": "country name from context",
  "ContextualQuestion": "Culturally adapted question text",
  "ContextualSolution": "Complete solution text in EXACT SAME FORMAT as input solution"
}

CRITICAL REQUIREMENTS:
- PHYSICAL PLAUSIBILITY IS MORE IMPORTANT THAN CULTURAL DIVERSITY
- If no culturally unique element fits physically, use a generic but realistic element
- Never sacrifice physics accuracy for cultural representation
- Return ONLY a valid JSON object with exactly these three fields
- Do NOT explain or describe what cultural terms are
\end{lstlisting}
\vspace{-0.25cm}
\caption{Prompt template used with Gemini 2.5 Pro for contextualizing physics problems}
\label{fig:adaptation_prompt}
\end{figure*}
\FloatBarrier


\newpage

\section{Model Inference and Evaluation Details}\label{inference_details}

\subsection{Parameters for cultural context generation}

For the cultural context database creation and contextualized question generation phases, we used slightly different parameters. Based on Google's recommendations for their models, we set temperature to 0.2 and top\_p to 0.95 when using Gemini 2.5 Pro for cultural database generation and contextual question generation. This slightly higher temperature value provided an appropriate balance between creativity and consistency, allowing for diverse cultural elements and problem formulations while maintaining coherence and factual accuracy. 

\subsection{Parameters for SLM inference}

For our primary experiments evaluating reasoning capabilities in SLMs, we used consistent inference parameters across all models to ensure fair comparison. All SLMs were run with a temperature of 0.1 and top\_p of 0.95. These low temperature settings were selected to minimize randomness and promote deterministic outputs, which is particularly important for assessing reasoning capabilities. Each model was configured with a maximum output length of 8,192 tokens to accommodate multi-step physics solutions. We additionally report the distribution of generated output lengths (in tokens) across models. 

\begin{figure*}[h]
\centering
\includegraphics[width=\textwidth]{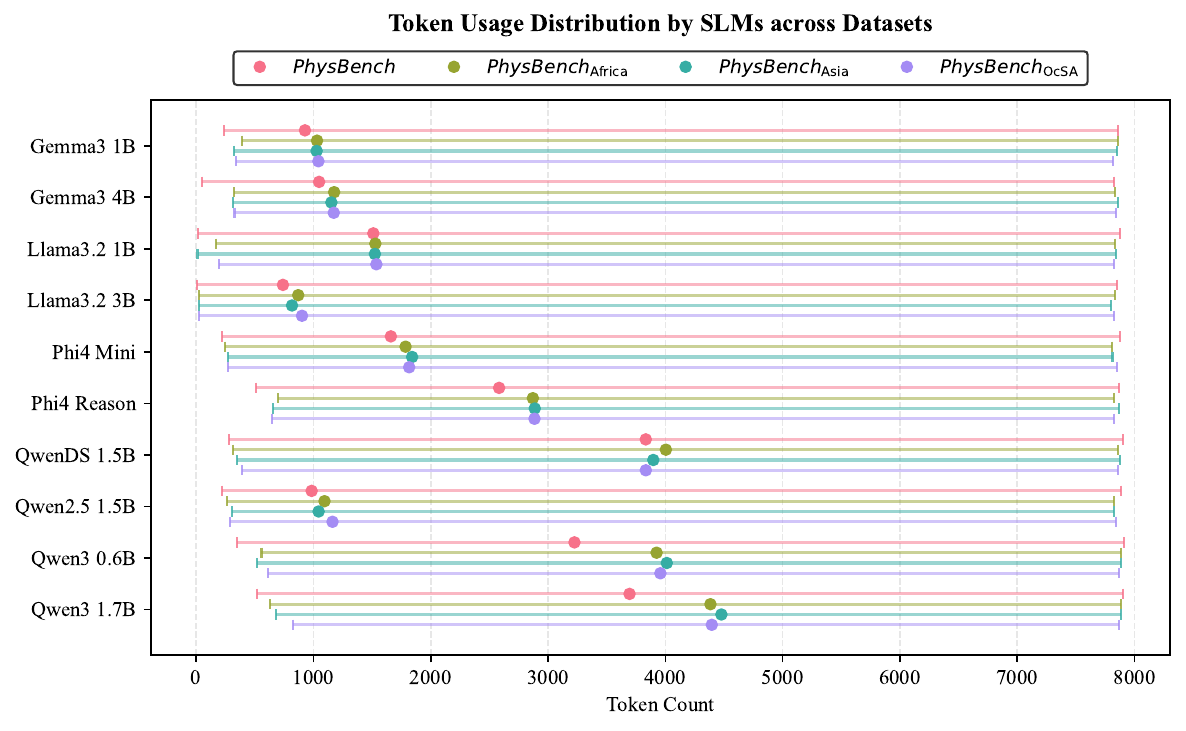}
\caption{Completion-token usage across models and dataset splits (\textit{PhysBench}, \textit{PhysBench}$_{\text{Asia}}$, \textit{PhysBench}$_{\text{Africa}}$, \textit{PhysBench}$_{\text{OcSA}}$). Bars show min--max completion tokens per solution and dots indicate medians (native tokenizer).}
\label{fig:token_distribution}
\end{figure*}

Figure~\ref{fig:token_distribution} summarizes completion-token usage for each SLM across multiple dataset splits (\textit{PhysBench}, \textit{PhysBench}$_{\text{Asia}}$, \textit{PhysBench}$_{\text{Africa}}$, \textit{PhysBench}$_{\text{OcSA}}$) under the same prompting and decoding configuration, with a fixed maximum generation length. Completion tokens are computed using each model’s native tokenizer. We observe substantial variance and long-tailed length distributions, reflecting model- and instance-dependent verbosity. Across models, the culturally contextualized variants tend to exhibit higher median completion lengths than \textit{PhysBench}, indicating that contextual rewrites can elicit longer solution traces under identical inference settings.

\subsection{Parameters for evaluation}

For our LLM-as-a-judge framework, we utilized Google's Gemini 2.5 Flash model with conservative sampling parameters (temperature = 0.1, top\_p = 0.95). These settings were selected to minimize stochasticity in the evaluation process, ensuring consistent and reliable assessment of both answer correctness and reasoning quality across all model outputs.

\subsection{Computational resources and cost analysis}

The creation of the culturally contextualized datasets was performed using Google's Gemini 2.5 Pro, incurring a total estimated cost of $\approx$350 USD, which includes token processing for approximately 50.5M input and 14.2M output tokens as well as Google Search grounding costs. The subsequent evaluation phase utilized Google's Gemini 2.5 Flash as the judge model, with a total cost of approximately 550 USD across all experimental runs (including ablations). Inference for the SLMs was conducted on in-house infrastructure using the vLLM library on NVIDIA RTX 6000 Pro and A6000 GPUs.

\section{Evaluation Rubric for Conceptual Questions}\label{eval_appendix}

The evaluation of conceptual questions utilizes a streamlined version of the P-REFS framework, adapted specifically for conceptual questions. Since these questions predominantly target the `Remember' and `Understand' cognitive levels, which generally require less multi-step reasoning" compared to problem-solving tasks, a simplified 5-point rubric was sufficient to capture the necessary reasoning dimensions without the granular complexity of the full framework.

\begin{table}[h]
\centering
\scriptsize
\caption{Evaluation framework for conceptual questions (5 points total).}
\label{tab:evaluation_criteria_conceptual}
\setlength{\tabcolsep}{4pt} 
\begin{tabular}{@{}ll@{}} 
\toprule
\textbf{Dimension} & \textbf{Evaluation Criteria}\\ 
\midrule
\multicolumn{2}{c}{\textbf{Part 1: Problem Formulation (1 point)}} \\
\midrule
Setup (0-1) & Correct identification of given information, approach,\\ 
& and assumptions.\\
\midrule
\multicolumn{2}{c}{\textbf{Part 2: Solution Execution (3 points)}} \\
\midrule
Conceptual (0-1) & Application of correct physics principles.\\
Execution (0-1) & Accuracy of formulas, calculations, and units.\\
 & \textit{(Marked N/A for purely conceptual questions)}\\
Coherence (0-1) & Logical flow and internal consistency of the solution.\\
\midrule
\multicolumn{2}{c}{\textbf{Part 3: Final Answer (1 point)}} \\
\midrule
Correctness (0-1) & Strict selection of the correct option.\\
 & \textit{(0.0 if ambiguous, multiple options, or incorrect)}\\
\bottomrule \\
\end{tabular}
\end{table}

The specific scoring criteria are detailed in Table~\ref{tab:evaluation_criteria_conceptual}. This rubric breaks down the solution into three critical phases: problem formulation, solution execution, and the final answer.

\section{Detailed Evaluation Protocol} \label{evaluation_protocol}

Our evaluation procedure is designed to assess reasoning quality rather than template compliance. When a model produces a structured response with clearly separated sections (\textit{Problem Formulation}, \textit{Step-by-step Solution}, \textit{Final Answer}), we evaluate each section independently using the corresponding P-REFS criteria, and the judge assigns criterion scores conditioned on the content in that section.

When a model does not provide clearly separable sections (i.e., the parsed output does not contain distinct fields for \texttt{Model\_ProblemFormulation}, \texttt{Model\_Step\_by\_step\_DetailedSolution}, and \texttt{Model\_FinalAnswer}), we apply a holistic evaluation mode. In this mode, the judge reads the entire response and assigns the same P-REFS criterion scores by locating the relevant content wherever it appears (even if steps are merged, unlabeled, written as continuous prose, or presented in a different order). The judge is explicitly instructed to score the presence and correctness of each P-REFS component based on substance, not formatting.

This adaptive evaluation ensures that (i) superficial formatting differences do not affect scores, (ii) all responses are evaluated under the same P-REFS rubric, and (iii) models with lower template compliance are assessed fairly on reasoning quality rather than instruction-following.

\section{LLM-as-a-Judge Validation} \label{judge_validation}

\subsection{Selection of judge models}

To identify the most reliable LLM-as-a-judge for automated P-REFS evaluation, we conducted a comparative validation study of three state-of-the-art large language models, with the objective of selecting a single judge model for final benchmark evaluation based on empirical validation metrics. The three candidate judges were selected based on their documented capabilities in mathematical and scientific reasoning tasks, architectural diversity to mitigate potential model-specific biases, and cost-efficiency for large-scale evaluation.

Gemini 2.5 Flash \citep{team2023gemini} represents one of the best Flash-class models with integrated thinking capabilities, specifically designed for advanced reasoning, coding, and mathematics while maintaining cost-efficiency and low latency. Qwen 3 32B is a dense 32.8 billion parameter model featuring hybrid reasoning with seamless switching between thinking mode (for complex logical reasoning, mathematics, and coding) and non-thinking mode (for efficient general-purpose dialogue), demonstrating strong performance on STEM benchmarks and multilingual support across 100+ languages~\citep{qwen3}. Llama 3.3 70B is a 70 billion parameter dense model that delivers performance comparable to Llama3.1 405B while being more than three times faster, achieving strong results on mathematical reasoning and coding benchmarks~\citep{llama3}.

We deliberately selected mid-to-large scale models rather than frontier-scale models (e.g., Gemini 2.5 Pro, OpenAI o3 or Claude Sonnet 3.5) for several reasons. First, our evaluation task of assessing whether SLM-generated physics solutions follow correct reasoning steps requires strong analytical capabilities but not the absolute frontier performance needed for novel problem solving or creative generation. Recent empirical studies have demonstrated that models in the 30B-70B parameter range can achieve strong performance as LLM judges, with Qwen 3 32B achieving state-of-the-art results on multiple judge benchmarks~\citep{j1reinforcement2025, codejudgebench2025}, while models like Llama 3 has demonstrated excellent alignment with human judges on structured evaluation tasks~\citep{thakur2024judging}. Second, practical deployment considerations: evaluating our complete benchmark requires processing tens of thousands of SLM responses across multiple criteria. The selected models offer competitive pricing through commercial APIs (Llama 3.3 70B via Together AI, Gemini 2.5 Flash via Google AI API) or can be deployed on local infrastructure (Qwen 3 32B on two NVIDIA RTX A6000 PRO GPUs). In comparison, frontier models cost substantially more per token, making large-scale assessment economically infeasible for research and educational applications. 

\subsection{Expert validation methodology}
To select the most reliable judge model for evaluating the complete \textit{PhysBench} benchmark, we conducted a comprehensive human validation study with two independent expert reviewers possessing graduate-level qualifications and domain expertise in physics.

\paragraph{Sampling Strategy:} We selected a stratified sample of question-response pairs to ensure comprehensive coverage across multiple dimensions: (1) all major physics topics (mechanics, thermodynamics, electromagnetism, optics, modern physics), and (2) all knowledge and skill categories defined in Bloom's taxonomy. For each of the 10 representative SLMs spanning different model families, parameter scales, and training approaches, we selected 55 questions, yielding 550 question-response pairs per judge model (55 questions $\times$ 10 SLMs). The specific questions varied across SLMs to ensure diverse coverage of physics reasoning patterns, with sampling designed to capture the full range of difficulty levels the judges must evaluate. All candidate judge models were evaluated on the same 550 question–response pairs, enabling a paired and controlled comparison across judges.

\begin{table*}[htbp]
\centering
\caption{Inter-rater reliability between two experts for different LLM-as-a-Judge variants across P-REFS criteria.}
\label{tab:interrater_comparison}
\begin{tabular}{l cc cc cc}
\toprule
 & \multicolumn{2}{c}{\textbf{Gemini 2.5 Flash}} & \multicolumn{2}{c}{\textbf{Qwen 3 32B}} & \multicolumn{2}{c}{\textbf{Llama 3.3 70B}} \\
\cmidrule(lr){2-3} \cmidrule(lr){4-5} \cmidrule(lr){6-7}
Criterion & Agr. (\%) & Gwet's AC2 & Agr. (\%) & Gwet's AC2 & Agr. (\%) & Gwet's AC2 \\
\midrule
\multicolumn{7}{c}{\textit{Problem Formulation}} \\
\midrule
Interpretation & 84.2 & 0.901 & 85.2 & 0.893 & 81.9 & 0.880 \\
Modeling       & 86.1 & 0.832 & 82.4 & 0.882 & 84.7 & 0.903 \\
\midrule
\multicolumn{7}{c}{\textit{Step-by-step Solution}} \\
\midrule
Relevance      & 87.0 & 0.852 & 81.5 & 0.881 & 81.9 & 0.784 \\
Application    & 85.2 & 0.908 & 85.2 & 0.908 & 75.0 & 0.680 \\
Calculation    & 83.3 & 0.803 & 85.2 & 0.894 & 86.1 & 0.906 \\
Correctness    & 87.0 & 0.852 & 87.0 & 0.918 & 77.8 & 0.723 \\
Coherence      & 88.9 & 0.876 & 79.6 & 0.861 & 84.7 & 0.822 \\
\midrule
\multicolumn{7}{c}{\textit{Final Answer}} \\
\midrule
Correctness    & 86.1 & 0.835 & 75.0 & 0.802 & 84.7 & 0.907 \\
\bottomrule
\end{tabular}
\end{table*}

\paragraph{Rating Protocol:} Each expert reviewer independently evaluated the judge assessments for the sampled question-response pairs across two distinct levels of granularity. To minimize potential bias, reviewers were blinded to the identity of the judge model (Gemini/Qwen/Llama) and evaluated anonymized judge outputs.

\begin{itemize} \item \textit{Criteria-level Assessment (Fine-grained)}: For each of the P-REFS criteria, reviewers rated their agreement with the judge's evaluation on a three-point ordinal scale: 2 (Full Agreement), indicating that the judge's score was entirely aligned with expert assessment; 1 (Partial Agreement), indicating that the judge's score was partially correct but lacked complete alignment with the expert's nuanced evaluation; and 0 (Disagree), indicating a fundamental misalignment or error in the judge's score. This granular approach allows for the pinpointing of specific reasoning dimensions where automated models may deviate from expert logic.

\item \textit{Question-level error classification (Aggregate)}: Upon completing the criteria-level review, experts assigned a singular behavioral label to the judge’s overall performance for that question. Judges were classified as -1 (Under-penalization/Lenient), 0 (Correct evaluation), or +1 (Over-penalization/Strict).

\end{itemize}

\paragraph{Inter-rater reliability analysis:} To assess the consistency of expert evaluations, we computed inter-rater reliability between the two independent reviewers. Although Cohen’s $\kappa$ is commonly used, it is sensitive to prevalence imbalance, the ``$\kappa$ paradox’’ \citep{feinstein1990high,byrt1993bias}, and can be artificially deflated when one rating category dominates. In our setting, the vast majority of criterion-level ratings are \textit{Full Agreement} (score 2), leading to high observed agreement but potentially low $\kappa$ values. We therefore report exact agreement percentages (on the three-level scale: 2 = Full, 1 = Partial, 0 = Disagree) and Gwet’s AC2 \citep{gwet2008computing}, which is more robust to prevalence effects. Since the rating scale is ordinal, we compute AC2 with linear ordinal weights over the three levels. Metrics are reported by P-REFS categories (Problem Formulation, Step-by-step Solution, Final Answer) to identify dimensions with higher or lower reviewer consistency.

Table~\ref{tab:interrater_comparison} summarizes inter-rater reliability across all P-REFS criteria for each candidate judge model. Exact agreement ranges from 75.0\% to 88.9\%, and Gwet’s AC2 ranges from 0.680 to 0.918, indicating consistently high reviewer agreement across criteria.

\subsection{LLM-as-a-Judge performance and error analysis}

\begin{table*}[htbp]
\centering
\caption{Expert-judge full-agreement rates across P-REFS criteria. Values represent the percentage of question-response pairs for which each expert rated the judge’s criterion-level assessment as \textit{Full Agreement} (score=2).}
\label{tab:full_agreement_details}
\begin{tabular}{l cc cc cc}
\toprule
 & \multicolumn{2}{c}{\textbf{Gemini 2.5 Flash}} & \multicolumn{2}{c}{\textbf{Qwen 3 32B}} & \multicolumn{2}{c}{\textbf{Llama 3.3 70B}} \\
\cmidrule(lr){2-3} \cmidrule(lr){4-5} \cmidrule(lr){6-7}
Criterion & Expert 1 & Expert 2 & Expert 1 & Expert 2 & Expert 1 & Expert 2 \\
\midrule
\multicolumn{7}{c}{\textit{Problem Formulation}} \\
\midrule
Interpretation  & 87.0 & 88.3 & 88.0 & 92.4 & 84.7 & 88.9 \\
Modeling        & 88.8 & 92.6 & 84.3 & 88.7 & 87.5 & 84.4 \\
\midrule
\multicolumn{7}{c}{\textit{Step-by-step Solution}} \\
\midrule
Relevance     & 91.7 & 94.4 & 83.3 & 88.1 & 84.7 & 87.2 \\
Application   & 88.0 & 91.5 & 85.2 & 86.3 & 80.6 & 84.4 \\
Calculation   & 92.6 & 90.7 & 88.0 & 86.3 & 88.9 & 87.2 \\
Correctness   & 93.4 & 91.7 & 91.7 & 90.4 & 83.3 & 84.4 \\
Coherence     & 94.1 & 92.7 & 84.3 & 83.5 & 93.1 & 91.7 \\
\midrule
\multicolumn{7}{c}{\textit{Final Answer}} \\
\midrule
Correctness   & 88.9 & 93.5 & 79.6 & 83.9 & 84.7 & 81.7 \\
\bottomrule
\end{tabular}
\end{table*}

Table~\ref{tab:full_agreement_details} reports criterion-level \textit{Full Agreement} rates between each expert reviewer and each candidate LLM judge. Across most criteria, Gemini 2.5 Flash shows consistently high full-agreement rates with both experts, particularly on step-by-step dimensions such as relevance, correctness, and coherence. Qwen 3 32B exhibits comparable full-agreement on several step-level criteria (e.g., step correctness and calculation), but shows notably lower full-agreement on final-answer correctness and coherence. Llama 3.3 70B is generally lower on several step-by-step criteria (e.g., application and step correctness) relative to Gemini 2.5 Flash.

\begin{figure}[h]
\centering
\includegraphics[width=\linewidth]{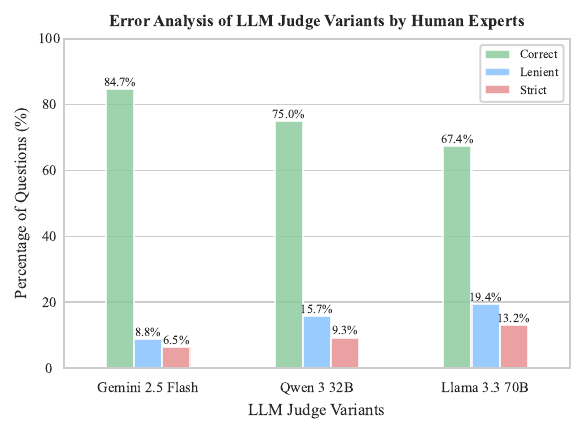}
\caption{Human expert error analysis of candidate LLM judges at the question level (Correct/Lenient/Strict).}
\label{fig:judge_error_analysis}
\end{figure}

In addition to criterion-level agreement, experts assigned an overall behavioral label per audited question: \textit{Correct} (0), \textit{Lenient} (-1; under-penalization), or \textit{Strict} (+1; over-penalization). Averaging across both experts, Gemini 2.5 Flash was labeled \textit{Correct} on 84.7\% of expert-evaluated questions, with 8.8\% \textit{Lenient} and 6.5\% \textit{Strict} cases; Qwen 3 32B achieved 75.0\% \textit{Correct} with 15.7\% \textit{Lenient} and 9.3\% \textit{Strict}; and Llama 3.3 70B achieved 67.4\% \textit{Correct} with 19.4\% \textit{Lenient} and 13.2\% \textit{Strict} (Figure~\ref{fig:judge_error_analysis}). Notably, all judges exhibit a consistent skew toward under-penalization (lenient $>$ strict), motivating the selection of an LLM judge that minimizes lenient failures, since these can mask grave reasoning errors in downstream benchmark statistics.

Based on these results, we selected Gemini 2.5 Flash as the LLM judge for evaluating the complete \textit{PhysBench} benchmark, given its superior correct evaluation rate (84.7\%), lowest error rates, and strong criterion-level alignment with expert assessments.

\section{Comprehensive Error Analysis using P-REFS}

\subsection{Error categories}\label{appen_error_analysis}

The P-REFS framework categorizes errors into eight distinct types, each corresponding to specific dimensions of the evaluation rubric. 

\paragraph{Interpretation Errors (INT):} These occur when models fail to correctly identify the given quantities, constraints, or what needs to be found in the problem. 

\textit{Example:} A problem states `A car accelerates from rest to 20 m/s in 5 seconds.' An interpretation error would be misidentifying the initial velocity as non-zero, or confusing the final velocity with acceleration.

\paragraph{Modeling Errors (MOD):} These arise when models select inappropriate solution strategies, make unjustified assumptions, or choose approaches that do not align with the problem requirements.

\textit{Example:} For a projectile motion problem, choosing to apply only vertical motion equations while ignoring horizontal components, or assuming negligible air resistance when the problem explicitly states it should be considered.

\paragraph{Conceptual Errors (CON):} These occur when models invoke incorrect physics principles or fail to identify the relevant physical concepts needed for the problem.

\textit{Example:} Using conservation of momentum instead of conservation of energy for an elastic collision problem, or applying Coulomb's law when the problem requires Ampere's law.

\paragraph{Concept Application Errors (APP):} These arise when models correctly identify the relevant physics concept but misuse it or apply the formula incorrectly.

\textit{Example:} Correctly identifying that Newton's second law ($F = ma$) is needed but then substituting mass for acceleration or vice versa, or using the kinetic energy formula $KE = \frac{1}{2}mv$ instead of $KE = \frac{1}{2}mv^2$.

\paragraph{Calculation Errors (CAL):} These are local arithmetic or algebraic manipulation errors (e.g., incorrect multiplication, sign errors, or incorrect rearrangement of an equation) assuming the underlying quantities and units are correctly specified.

\textit{Example:} Computing $3 \times 4 = 16$ instead of $12$, or incorrectly simplifying $\frac{10}{2} + 5$ as $5 + 5 = 15$ instead of $5 + 5 = 10$. This category also includes errors in algebraic rearrangement, such as solving $2x + 3 = 11$ and obtaining $x = 3$ instead of $x = 4$.

\paragraph{Correctness Errors (COR):} These occur when intermediate or final quantities are physically inconsistent, including unit/dimensional mismatches, incorrect unit conversions, or carrying forward an inconsistent intermediate value to later steps. Unlike CAL, which detects local arithmetic or algebraic errors, COR identifies failures in unit/dimension validity, as well as the propagation/validation of intermediate results across steps.

\textit{Example:} Mixing units (adding meters to centimeters without conversion), reporting acceleration with units of velocity (m/s instead of m/s$^2$), or carrying forward an incorrect intermediate result to subsequent steps even if the calculation method is correct.

\paragraph{Coherence Errors (COH):} These occur when logical flow is disrupted through illogical step transitions, skipped reasoning, circular logic, or contradictory statements within the solution.

\textit{Example:} Jumping from initial conditions directly to a final answer without showing intermediate steps, using a result before deriving it, or making statements that contradict earlier parts of the solution (e.g., stating `velocity is constant' in one step but then calculating acceleration in the next).

\paragraph{Final Answer Errors (ANS):} These represent incorrect numerical results, missing or wrong units in the final answer, or (for multiple-choice questions) selecting an incorrect option.

\textit{Example:} Computing the correct intermediate steps but making an error in the final calculation, reporting an answer as `50 m' when the correct answer is `50 m/s,' or selecting option (B) when the correct answer is (C).

\subsection{Error type prevalence analysis} \label{appen_error_prevalence}

While Figure~\ref{fig:error_survival} reveals where reasoning breaks down sequentially, this section examines the independent frequency of each error type. Errors are categorized into eight types aligned with the rubric: Interpretation, Modeling, Conceptual, Concept Application, Calculation, Correctness, Coherence, and Final Answer. A single response may exhibit multiple error types; therefore, categories are not mutually exclusive and the reported frequencies should be interpreted independently.

\begin{figure*}[h!]
\centering
\includegraphics[width=0.9\textwidth]{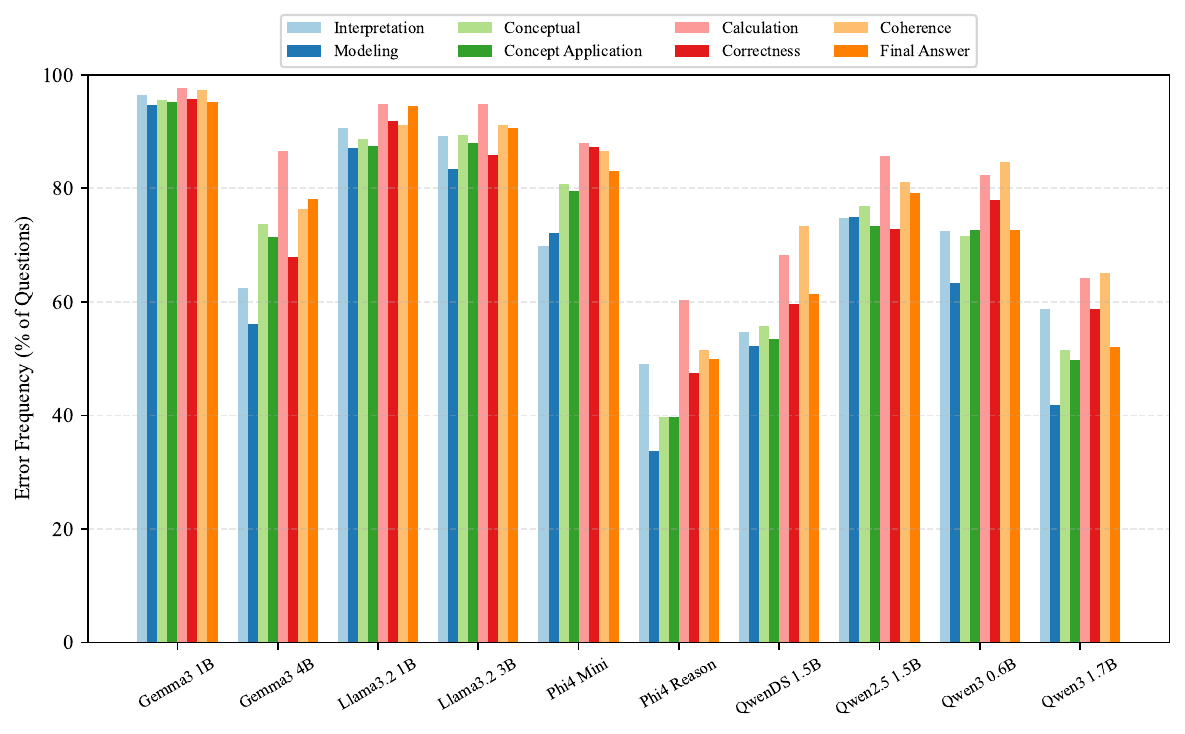}
\caption{Error frequencies across P-REFS evaluation dimensions. Each bar represents the percentage of problem-solving questions exhibiting each specific error type. Categories are not mutually exclusive.}
\label{fig:error_frequency}
\end{figure*}

Figure~\ref{fig:error_frequency} reports the prevalence of each error type across the evaluated SLMs. Two high-level patterns emerge.

\paragraph{Errors are multi-factorial rather than single-point failures.}
Across all models, no single error type exclusively dominates. Instead, failures are distributed across early-stage formulation and late-stage execution. For example, even the strongest models like Phi4 Reason and Qwen3 1.7B show non-trivial error rates in both Modeling ($\approx$34\%) and Calculation ($\approx$60\%), indicating that improving physics reasoning requires end-to-end improvements spanning problem understanding, model selection, and procedural execution, rather than addressing a single isolated weakness.

\paragraph{Early-stage vs. late-stage errors separate weaker and stronger models:}
Weaker models exhibit substantially higher prevalence of early-stage errors (INT, MOD), indicating that many solutions fail before reaching coherent execution. For instance, Gemma3 1B shows near-saturation at the earliest stages (INT $\approx$96\%, MOD $\approx$95\%), implying that incorrect problem reading or setup occurs for almost all items. This aligns with the stage-wise degradation trends in the main paper, where weaker models collapse immediately during comprehension or initial formulation. Mid-tier models reduce these early-stage failures (e.g., INT on the order of $\approx60-75\%$), but still exhibit high execution-related error prevalence (CAL/COR often $\approx70-90\%$) and frequent final-answer failures (ANS $\approx60-80\%$). This indicates that while mid-tier models more often identify relevant quantities and choose partially appropriate approaches, they struggle to maintain procedural correctness throughout multi-step derivations.

\paragraph{Execution reliability remains a persistent bottleneck:}
Even among stronger models, execution-related error types remain prominent. For example, Phi4 Reason still exhibits calculation errors at $\approx$60\% and correctness errors at $\approx$56\%, while Qwen3 1.7B shows similarly high prevalence (CAL $\approx$64\%, COR $\approx$59\%). These CAL/COR failures reflect fragility in numerical manipulation, unit handling, and intermediate-result validation, and help explain why correct final answers are often not supported by fully correct end-to-end solutions: errors can emerge during intermediate computation even when the high-level approach is appropriate.

\paragraph{Coherence errors occur less frequently but reflect compounding failures:}
Coherence errors are comparatively less prevalent than the INT/MOD and CAL/COR in many models, suggesting that when the underlying setup and computations remain consistent, models often produce logically connected narratives. However, coherence failures still appear when earlier mistakes compound, leading to contradictions or skipped reasoning that prevents recovery.

\subsection{Qualitative failure modes observed in expert validation:}
In addition to the quantitative prevalence trends in Figure~\ref{fig:error_frequency}, our expert validation of the LLM-as-a-judge outputs surfaced several recurring qualitative failure modes that help interpret the rubric-aligned error frequencies.

\paragraph{Unit-only partial credit despite incorrect final selection:}
Experts observed cases where models produced a unit expression consistent with the target quantity while selecting an incorrect multiple-choice option or reporting an incorrect magnitude. In such cases, judges may still award partial credit on the unit dimension despite an incorrect final answer.

\paragraph{``Thinking spirals'' and coherence degradation at higher-order questions:}
Experts also noted ``thinking spirals''--long, unstable reasoning traces that branch or repeatedly revise intermediate steps without converging to a final answer. These spirals were more common on higher-Bloom items and typically coincided with lower coherence due to contradictions, missing links, or unresolved intermediate quantities.

\paragraph{Correct answers via imperfect reasoning and option rounding:}
Experts found instances where models reached the correct multiple-choice option despite an invalid or incomplete method, often because an approximate numeric estimate rounded to (or was nearest to) the correct option. This pattern helps explain why answer-correct outputs can still exhibit substantive execution or correctness errors.

Together, these observations complement the quantitative breakdown in Figure~\ref{fig:error_frequency} by illustrating how errors can manifest in practice (e.g., separating unit correctness from value correctness, or coherence failures driven by prolonged unstable reasoning), and they motivate the rubric’s separation of formulation, execution, and final-answer dimensions.

\section{Effect of Template Compliance in Performance}\label{template_compliance}

We analyzed template compliance across all model responses for problem-solving questions. A response is classified as `template-compliant' if the parsed JSON contains distinct fields for \texttt{Model\_ProblemFormulation}, \texttt{Model\_Step\_by\_step\_DetailedSolution}, and \texttt{Model\_FinalAnswer}. Non-compliant responses include outputs where these components are present but unlabeled, merged into continuous prose, or presented in a different order. Importantly, non-compliant responses are not discarded: they are evaluated using the same P-REFS rubric in a single inference step that scores each rubric component based on content, independent of formatting.

Table~\ref{tab:template_compliance} reports per-model compliance rates and average P-REFS scores split by compliant vs.\ non-compliant responses.

\begin{table*}[h]
\centering
\footnotesize
\caption{Template compliance rates and performance by model for problem-solving questions.}
\label{tab:template_compliance}
\begin{tabular}{lcccc}
\toprule
Model & Compliance Rate & Avg P-REFS & Avg P-REFS  & $\Delta$ Score \\
&  (\%) &  (Compliant) & (Non-compliant) & \\
\midrule
Gemma3 1B     & 83.6 & 1.38 & 1.37 & +0.00 \\
Gemma3 4B     & 96.9 & 4.93 & 2.71 & +2.22 \\
Llama3.2 1B   & 65.4 & 1.68 & 1.65 & +0.03 \\
Llama3.2 3B   & 63.2 & 3.61 & 0.21 & +3.40 \\
Phi4 Mini     & 64.9 & 3.92 & 2.18 & +1.74 \\
Phi4 Reason   & 70.3 & 7.57 & 5.44 & +2.13 \\
QwenDS 1.5B   & 43.4 & 6.28 & 5.36 & +0.92 \\
Qwen2.5 1.5B  & 93.7 & 4.31 & 3.05 & +1.25 \\
Qwen3 0.6B    & 59.6 & 3.76 & 5.03 & -1.27 \\
Qwen3 1.7B    & 92.0 & 6.78 & 3.66 & +3.12 \\
\bottomrule
\end{tabular}
\end{table*}

\begin{figure}[h]
\centering
\includegraphics[width=\linewidth]{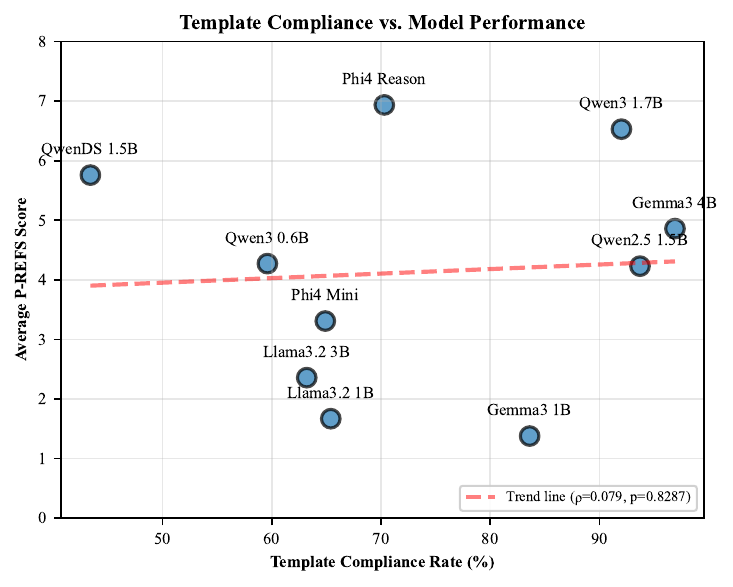}
\caption{Template compliance rate versus average P-REFS score across 10 SLMs. The near-flat trend line ($\rho$ = 0.079, $p$ = 0.83) demonstrates statistical independence between instruction-following and reasoning capability.}
\label{fig:compliance_performance}
\end{figure}

\paragraph{Compliance and reasoning performance are weakly coupled:}
Across the 10 SLMs, template compliance shows no significant correlation with overall reasoning performance (Spearman's $\rho=0.079$, $p=0.83$), indicating that instruction-following (format adherence) and physics reasoning are largely independent competencies. Figure~\ref{fig:compliance_performance} visualizes this dissociation: for example, QwenDS~1.5B attains strong average P-REFS despite the lowest compliance rate, whereas Gemma3~1B exhibits relatively high compliance but low P-REFS scores.

\paragraph{Compliance varies substantially across models:} Compliance rates span a wide range (43.4\%--96.9\%; Table~\ref{tab:template_compliance}). Larger or better instruction-tuned models tend to be more compliant (e.g., Gemma3~4B, Qwen2.5~1.5B, Qwen3~1.7B), while smaller/weaker models exhibit more structural variation (e.g., QwenDS~1.5B, Qwen3~0.6B, Llama3.2~3B). In addition, for reasoning-oriented models, a subset of non-compliant outputs reflects long reasoning traces that fail to converge to a clearly stated final answer, producing continuous prose and occasionally omitting an explicit \textit{Final Answer}; our parsing failure evaluation mode is designed to score these responses based on content while appropriately reflecting missing or ambiguous final conclusions under the relevant P-REFS dimensions.

\paragraph{Adaptive evaluation reduces structural bias:} Within each model, the score gap between compliant and non-compliant responses is not uniform (Table~\ref{tab:template_compliance}): most models show modest positive differences, while at least one model (Qwen3~0.6B) performs slightly better in free-form mode ($\Delta=-1.27$). Overall, the magnitude of this gap is moderate (mean $|\Delta|=1.61$, median $|\Delta|=1.50$ on a 10-point scale), and its direction is inconsistent across models. This supports our design choice to score non-compliant outputs holistically under the same P-REFS criteria, ensuring that models are compared primarily on reasoning quality rather than on template adherence.

\section{Formulation-scaffolded Inference Ablation}\label{with_problemformulation}

To diagnose whether problem-solving failures arise primarily from early-stage setup versus downstream execution, we conduct a formulation-scaffolded ablation in which we prepend the reference Problem Formulation (from our structured solutions) to the model prompt. Models are then asked to generate the remaining sections (Step-by-step Solution and Final Answer) under the same decoding configuration as in the main experiments. We evaluate these outputs using the same pipeline and report (i) \textsc{Final Answer Correct} and (ii) \textsc{Fully Correct}, where the latter requires Parts 2 and 3 to be fully correct while treating Part 1 as satisfied.

\begin{figure*}[t]
\centering
\includegraphics[width=0.8\textwidth]{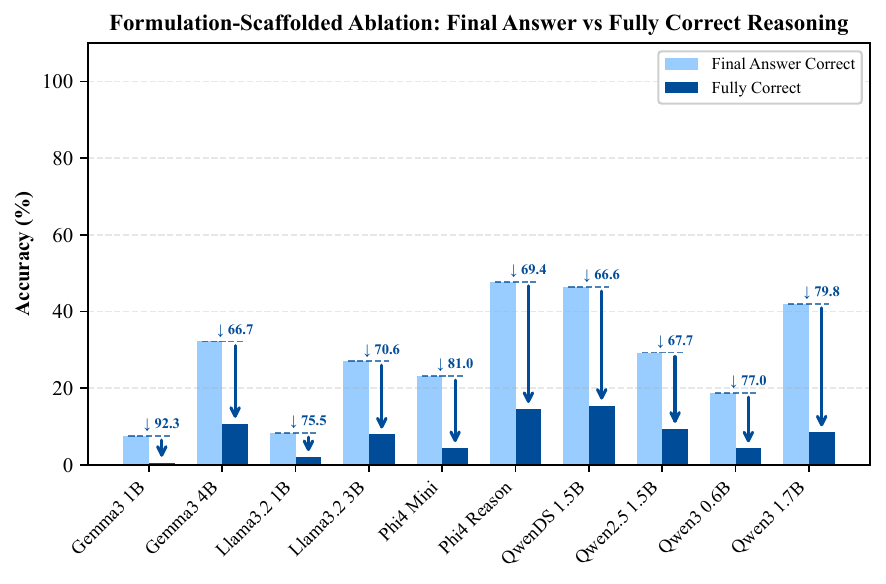}
\caption{Formulation-scaffolded ablation on problem-solving questions. We prepend the reference \textit{Problem Formulation} to the prompt and score \textsc{Final Answer Correct} versus \textsc{Fully Correct} (i.e., Parts 2 and 3 fully correct). The annotated arrows indicate the relative gap between final-answer accuracy and fully-correct reasoning under this setting.}
\label{fig:formulation_ablation}
\end{figure*}

Figure~\ref{fig:formulation_ablation} compares the final-answer accuracy and the fully-correct (P2+P3) accuracy under this scaffolded setting, highlighting how much of the score gap persists even when formulation is provided.

Across the 10 SLMs, providing the formulation increases average final-answer accuracy from 23.15\% to 28.26\% (+5.11 points) and increases average fully-correct accuracy from 3.32\% to 7.81\% (+4.48 points). Improvements in \textsc{Final Answer Correct} are largest for Llama3.2~3B (+18.58 points; 8.49\%$\rightarrow$27.07\%) and Gemma3~4B (+11.74 points; 20.46\%$\rightarrow$32.20\%). \textsc{Fully Correct} increases for every model, with the largest gains for QwenDS~1.5B (+8.34 points; 7.14\%$\rightarrow$15.48\%), Gemma3~4B (+8.18 points; 2.55\%$\rightarrow$10.73\%), Qwen2.5~1.5B (+7.99 points; 1.47\%$\rightarrow$9.46\%), and Llama3.2~3B (+7.76 points; 0.19\%$\rightarrow$7.95\%). 

Three models show reduced \textsc{Final Answer Correct} under formulation scaffolding (Phi4~Reason: 49.00\%$\rightarrow$47.76\%; Qwen3~0.6B: 26.14\%$\rightarrow$18.76\%; Qwen3~1.7B: 47.37\%$\rightarrow$42.01\%), while still improving in \textsc{Fully Correct}, indicating that providing formulation can alter how reasoning-specialized models 
structure their outputs. We hypothesize that this pattern reflects a generation-structure mismatch in reasoning-specialized models. When the formulation is provided externally, these models may change how they organize the remainder of the solution (e.g., producing longer traces, more backtracking, or delayed commitment), which can reduce the probability of ending with a correct final answer. 

\begin{figure*}[t]
    \centering
    \includegraphics[width=\textwidth]{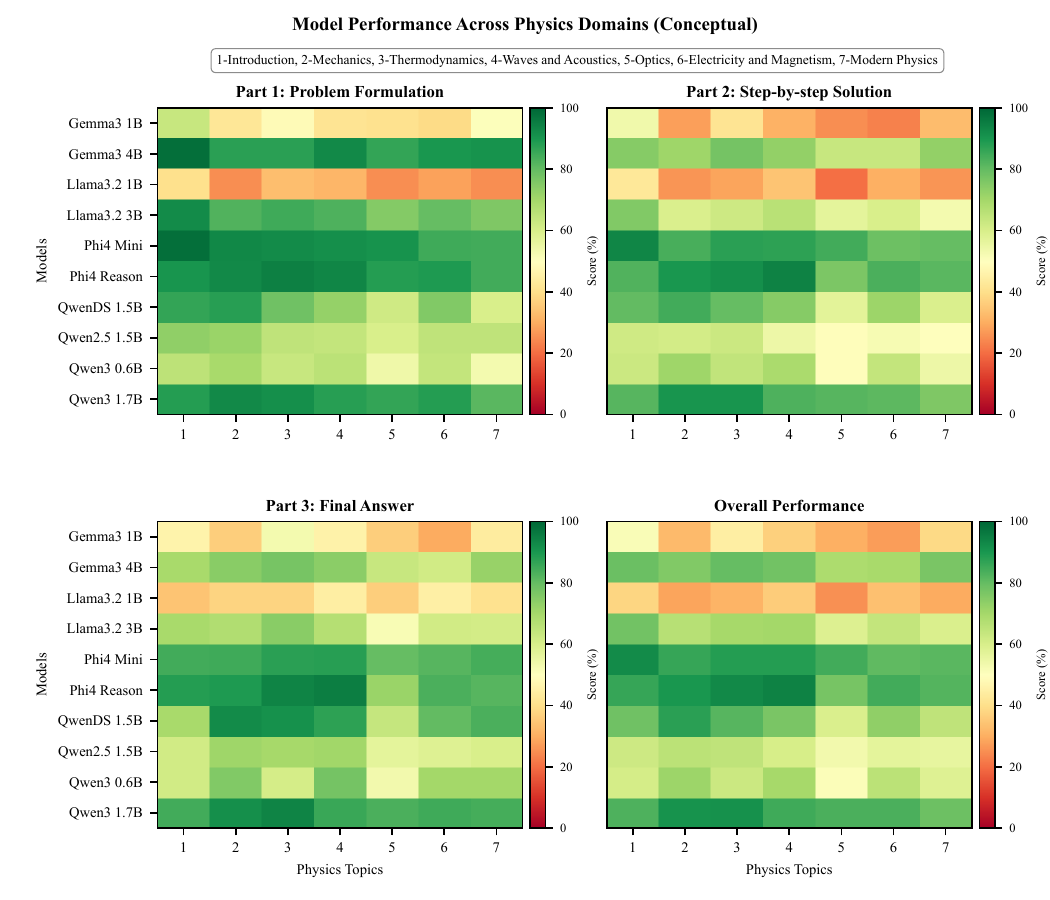}
    \caption{Performance across physics domains for conceptual questions. Four-panel heatmap showing Part 1 (Problem Formulation), Part 2 (Step-by-step Solution), Part 3 (Final Answer), and Overall Performance for 10 SLMs across 7 physics domains: Introduction, Mechanics, Thermodynamics, Waves and Acoustics, Optics, Electricity and Magnetism, and Modern Physics. Color intensity represents performance level (green = higher, red = lower scores).}
    \label{fig:physics_topics_conceptual}
\end{figure*}

\begin{figure*}[t]
    \centering
    \includegraphics[width=\textwidth]{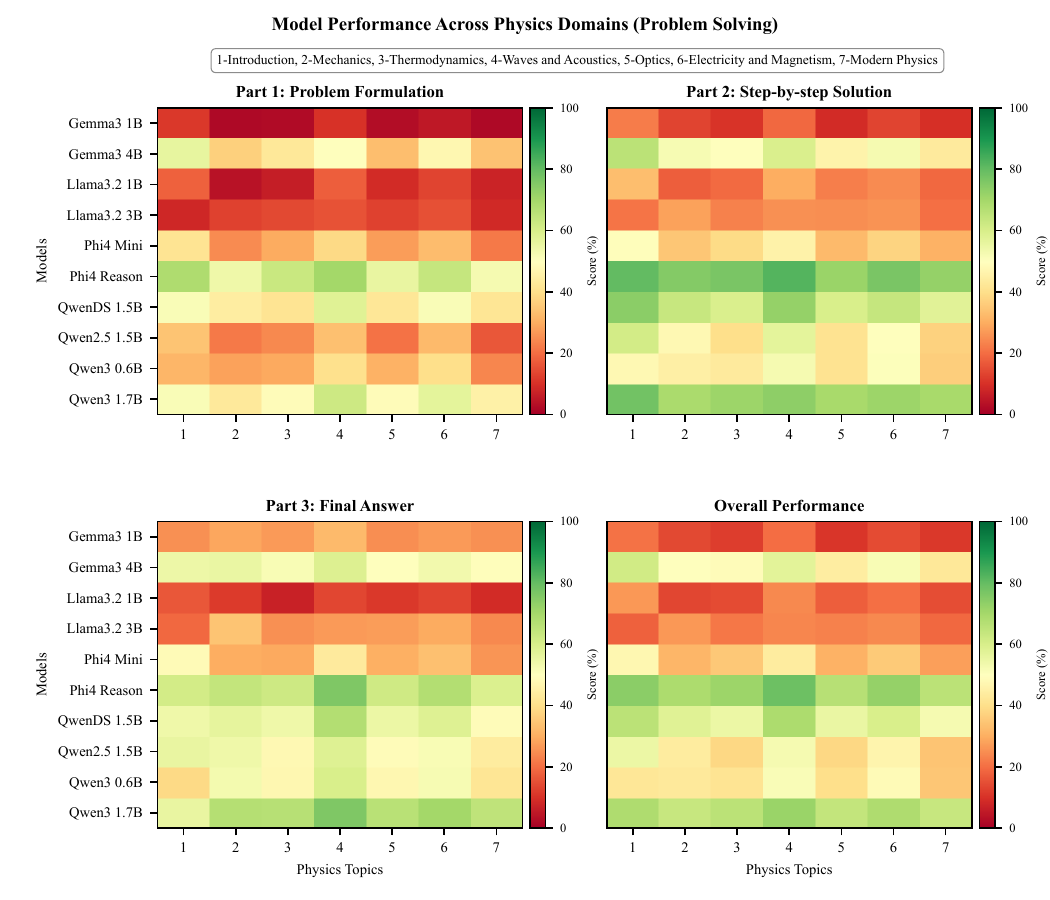}
    \caption{Performance across physics domains for problem-solving questions. Four-panel heatmap showing Part 1 (Problem Formulation), Part 2 (Step-by-step Solution), Part 3 (Final Answer), and Overall Performance for 10 SLMs across 7 physics domains: Introduction, Mechanics, Thermodynamics, Waves and Acoustics, Optics, Electricity and Magnetism, and Modern Physics. Color intensity represents performance level (green = higher, red = lower scores).}
    \label{fig:physics_topics_problemsolving}
\end{figure*}

\begin{figure*}[t]
    \centering
    \includegraphics[width=\textwidth]{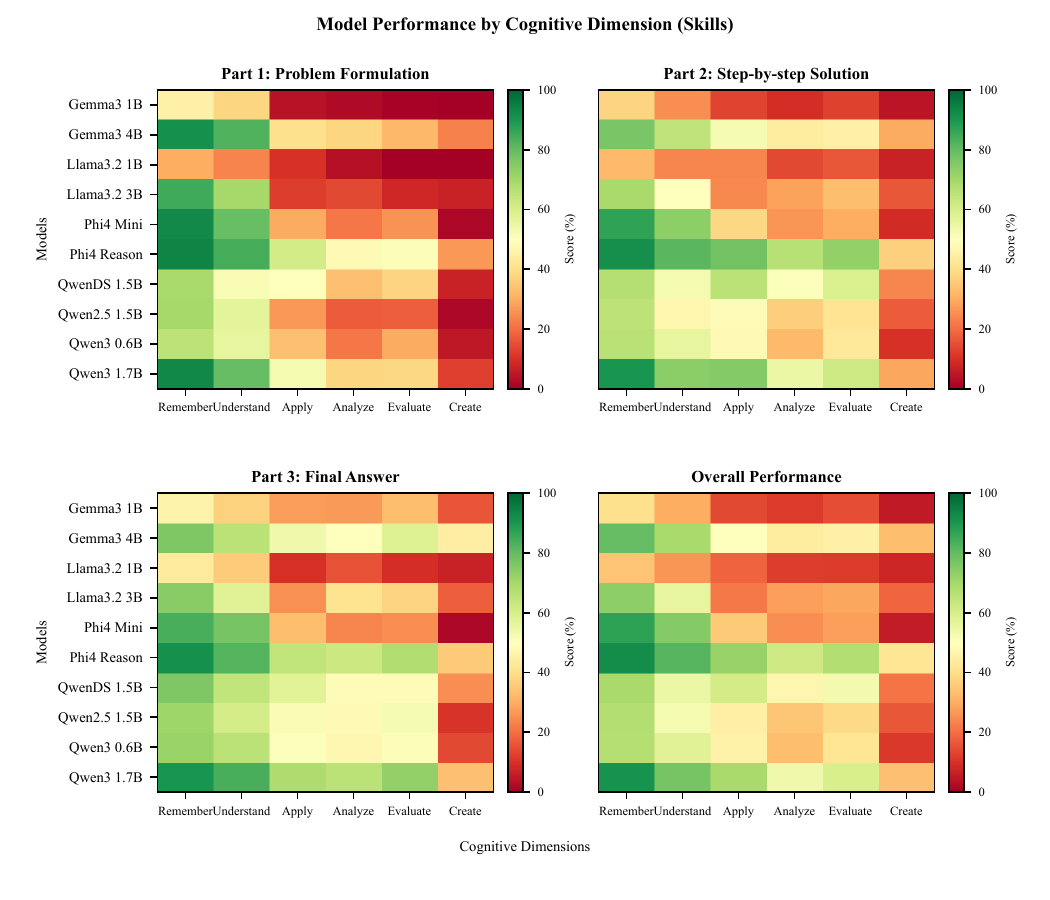}
    \caption{Performance across Bloom's Taxonomy cognitive dimensions. Four-panel heatmap showing Part 1 (Problem Formulation), Part 2 (Step-by-step Solution), Part 3 (Final Answer), and Overall Performance for 10 SLMs across six cognitive levels: Remember, Understand, Apply, Analyze, Evaluate, and Create. Color intensity represents performance level (green = higher, red = lower scores). A clear degradation pattern emerges as cognitive complexity increases, with models demonstrating strong performance on lower-order skills (Remember, Understand) but substantial deterioration on higher-order thinking (Analyze, Evaluate, Create).}
    \label{fig:cognitive_dimensions}
\end{figure*}

\begin{figure*}[t]
    \centering
    \includegraphics[width=\textwidth]{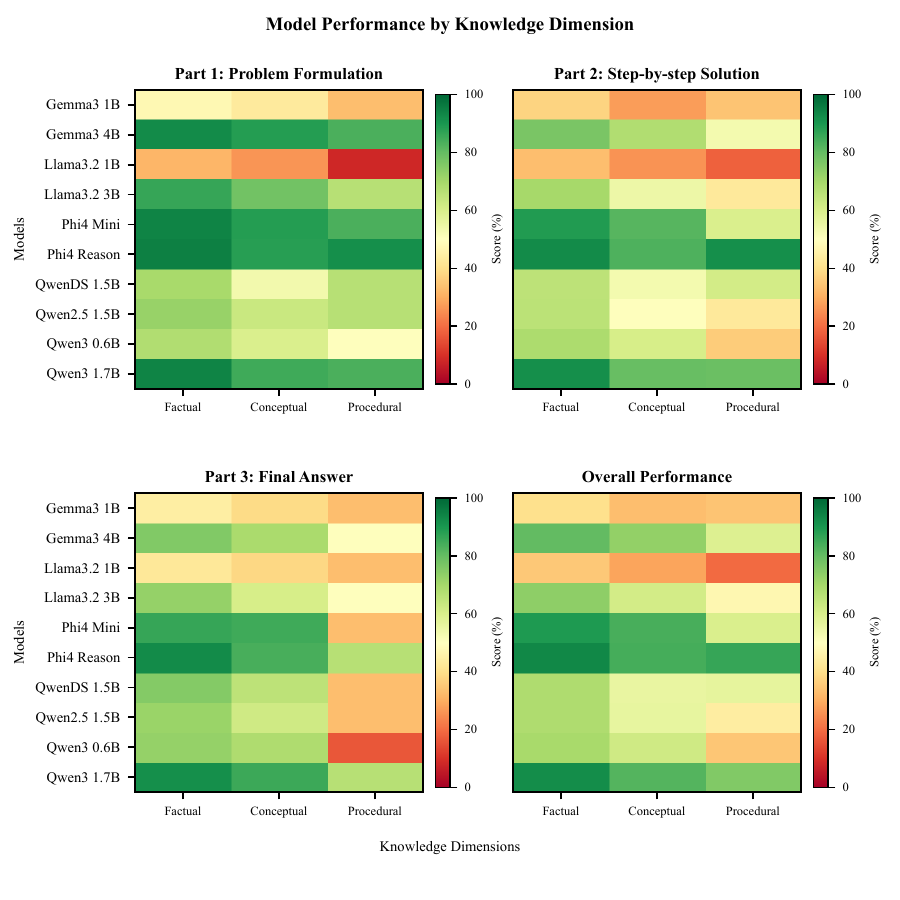}
    \caption{Performance across Bloom's Taxonomy knowledge dimensions. Four-panel heatmap showing Part 1 (Problem Formulation), Part 2 (Step-by-step Solution), Part 3 (Final Answer), and Overall Performance for 10 SLMs across three knowledge types: Factual (recall of terminology and basic elements), Conceptual (understanding of relationships and structures), and Procedural (application of methods and techniques). Color intensity represents performance level (green = higher, red = lower scores). Models demonstrate strongest performance on Factual knowledge, moderate performance on Conceptual knowledge, and weakest performance on Procedural knowledge, indicating difficulty with systematic application of physics methods.}
    \label{fig:knowledge_dimensions}
\end{figure*}

\end{document}